\documentclass[runningheads]{llncs}
\usepackage{graphicx}

\usepackage{tikz}
\usepackage{comment}
\usepackage{amsmath,amssymb} %
\usepackage{color, colortbl}
 \usepackage{multirow}
\usepackage{xcolor,pifont}
\usepackage{booktabs}

\definecolor{Gray}{gray}{0.9}

\usepackage[accsupp]{axessibility}  %

\usepackage[width=122mm,left=12mm,paperwidth=146mm,height=193mm,top=12mm,paperheight=217mm]{geometry}

\usepackage{listings}
\usepackage{algorithm}
\usepackage{algorithmic}

\begin{document}
\pagestyle{headings}
\mainmatter
\def\ECCVSubNumber{3550}  %

\title{Disentangled Representation Learning for Text-Video Retrieval} %

\titlerunning{Disentangled Representation Learning for Text-Video Retrieval}
\author{Qiang Wang, Yanhao Zhang, Yun Zheng, Pan Pan, Xian-Sheng Hua\\
\institute{DAMO Academy, Alibaba Group\\
{\email \{qishi.wq, yanhao.zyh, zhengyun.zy, panpan.pp, xiansheng.hxs\}@alibaba-inc.com}}
}
\maketitle

\begin{abstract}
Cross-modality interaction is a critical component in Text-Video Retrieval (TVR), yet there has been little examination of how different influencing factors for computing interaction affect performance.
This paper first studies the interaction paradigm in depth, where we find that its computation can be split into two terms, the interaction contents at different granularity and the matching function to distinguish pairs with the same semantics.
We also observe that the single-vector representation and implicit intensive function substantially hinder the optimization.
Based on these findings, we propose a disentangled framework to capture a sequential and hierarchical representation. 
Firstly, considering the natural sequential structure in both text and video inputs, a Weighted Token-wise Interaction (WTI) module is performed to decouple the content and adaptively exploit the pair-wise correlations.
This interaction can form a better disentangled manifold for sequential inputs.
Secondly, we introduce a Channel DeCorrelation Regularization (CDCR) to minimize the redundancy between the components of the compared vectors, which facilitate learning a hierarchical representation.
We demonstrate the effectiveness of the disentangled representation on various benchmarks, \textit{e.g.}, surpassing CLIP4Clip largely by $+2.9\%$, $+3.1\%$, $+7.9\%$, $+2.3\%$, $+2.8\%$ and $+6.5\%$ R@1 on the MSR-VTT, MSVD, VATEX, LSMDC, AcitivityNet, and DiDeMo, respectively.

\end{abstract}

\section{Introduction}

\begin{figure}[t]
\centering
\includegraphics[width=0.95\linewidth]{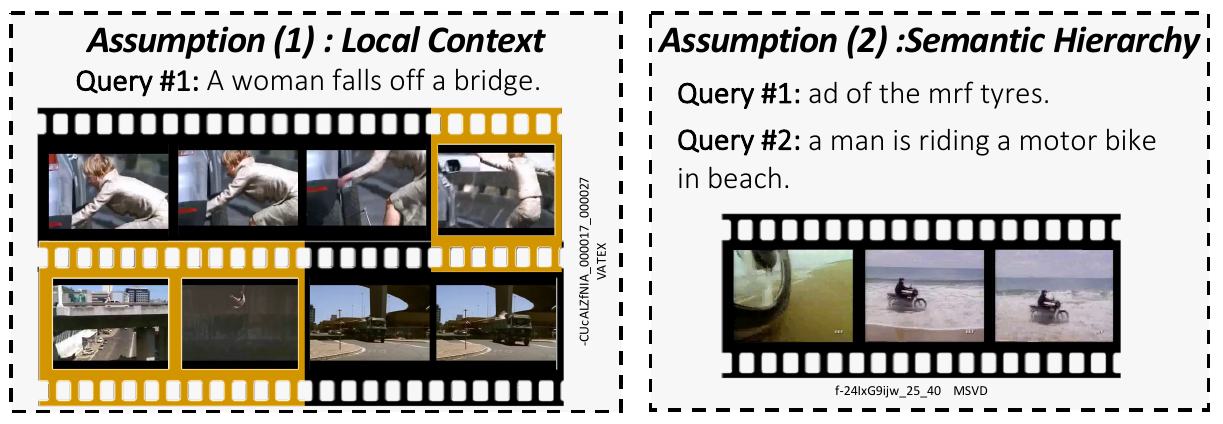}
\vspace{-3mm}
\caption{\textbf{Illustrations of common underlying assumptions for Text-Video Retrieval.}  (1) The description could be relevant to the local video context. (2) Multiple distinct sentences can hint at the same video. Our disentangled representation addresses both assumptions through token-wise interaction and channel decorrelation. }
\vspace{-2mm}
\label{fig:intro}
\end{figure}

Text-Video Retrieval (TVR) has significant worth with the explosive growth of video content on the internet. 
Alignment between different modalities requires particular consideration of both intra-modal representation and cross-modality interaction.
The inherent asymmetry across different modalities raises an expression challenge for Text-Video Retrieval.
The pioneering works~\cite{mmt,ce} incorporate multi-modality features and aggregate information from different pre-trained experts to boost performance.
While the open-ended text queries and diverse vision contents require tremendous labor. 

With the great success of NLP pre-training~\cite{bert,roberta,bart,gpt,t5}, Vision-Language Pre-Training (VLPT) has received increasing attention~\cite{hero,clipbert,oscar,univl,frozen}.
Recently, the Contrastive Language–Image Pre-training (CLIP~\cite{clip}) leverages a massive amount of image-caption pairs to learn generic vision and language representations, showing impressive improvements in diverse vision-language (VL) tasks.
Researchers~\cite{sclip,frozen,clip4clip,clip2video,clip2tv} explore the power of this design principle to text-video retrieval with additional temporal fusion module.
CLIP4Clip~\cite{clip4clip} utilizes a temporal transformer to aggregate sequential features into a \textit{single} high-dimensional representation, and retrievals by dot-product search. 

However, the text-video retrieval is uniquely characterized by its sequential frames.
Thus we highlight the common underlying  assumptions behind TVR in Figure~\ref{fig:intro}:
(1) the description related with the local segment still needs to be retrieved, and
(2) human-generated sentences naturally have a hierarchical structure and can describe in different views.
Based on these perspectives, we can re-examine the existing algorithmic framework, especially the \textbf{interaction} components.

\begin{figure}[t]
\centering
\includegraphics[width=0.99\linewidth]{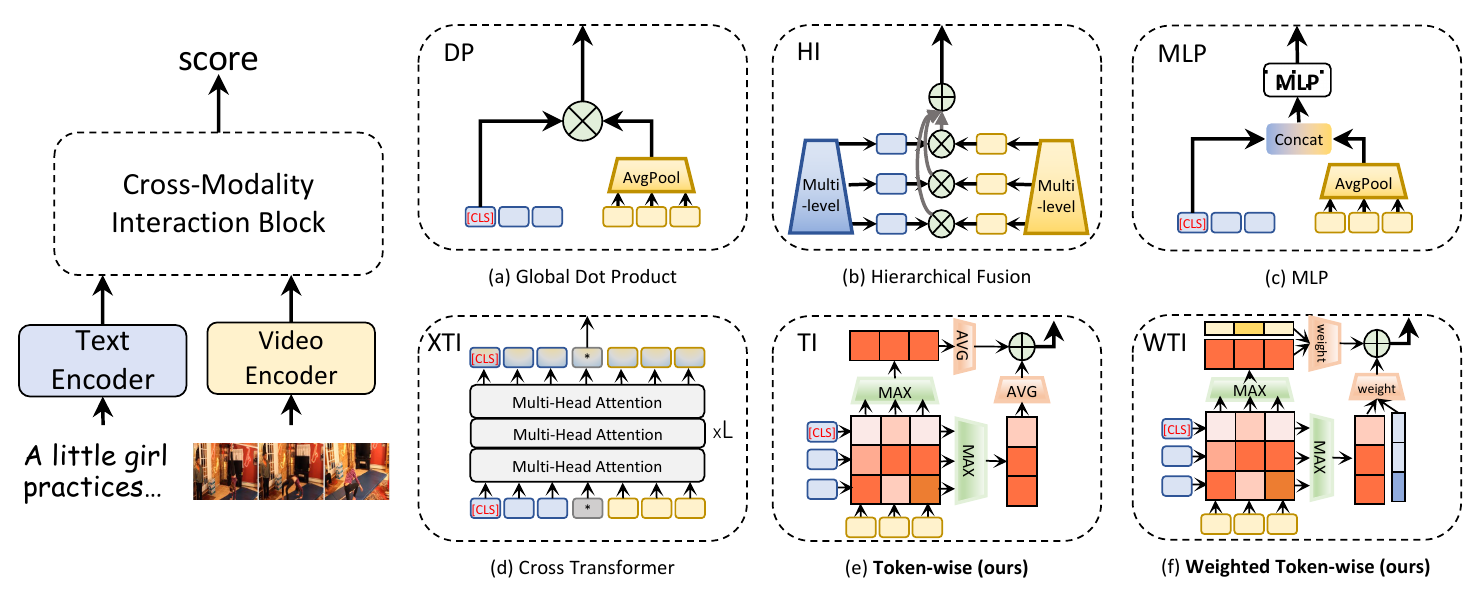}
\vspace{-2mm}
\caption{\textbf{Illustrations of the text-video retrieval architecture and six categories of interaction methods.} Each subplot shows a typical interaction module.}
\label{fig:arch}
\end{figure}

Figure~\ref{fig:arch} shows typical interaction archetypes and the proposed variations.
In determining the process flow for the interaction block, only a few properties are commonly considered. 
One is the granularity of input content, and the other is the interaction function.

The single-vector representation is widely used in the fields of biometrics~\cite{arcface,reid} and text retrieval~\cite{dssm}, and its retrieval process is extremely concise and efficient.
While as shown in Figure~\ref{fig:arch} (a), the over-abstract representation will introduce a lack of fine-grained matching capabilities.
Therefore, the Mixtures of Experts (MoE) approaches~\cite{ce,mmt,hit} take advantage of individual domains to integrate generalizable aggregated features.
MMT~\cite{mmt} and CE~\cite{ce} explicitly construct a multi-expert fusion mechanism to boost retrieval performance.
In addition, HIT~\cite{hit} performs hierarchical cross-modality contrastive matching at both feature-level and semantic-level. They can all be summed up as a hierarchical fusion architecture (Figure~\ref{fig:arch}(b)).

In contrast to the above parameter-free dot-product function, the deeply-contextualized interaction has emerged that fine-tunes MLP models for estimating relevance~\cite{spm,jsfusion} (Figure~\ref{fig:arch}(c)).
The recent cross transformer interaction approaches~\cite{hero,clip4clip} can inference the complex relationship between \textit{arbitrary}-length text and video. 
However, the neural network interaction lacks the typical inductive bias for matching and usually suffers from optimization difficulty and performance degradation~\cite{clip4clip}.
Furthermore, these heavy interactions will bring a prohibitively computational cost for real-world deployments.

We present a disentangled framework to address the above challenges, where a novel token-wise interaction and channel decorrelation regularization collaboratively decouples the sequential and hierarchical representation. 
Concretely, we propose a lightweight token-wise interaction that fully interacts with all sentence tokens and video frame tokens (Figure~\ref{fig:arch}(e-f)). 
Compared with the single-vector interaction (Figure~\ref{fig:arch}(a)(c)) and multi-level interaction (Figure~\ref{fig:arch}(b)), our method can preserve more fine-grained clues.
Compared to the cross transformer interaction (Figure~\ref{fig:arch}(d)), the proposed interaction mechanism significantly ease the optimization difficulty and computational overhead.
In addition to the interaction mechanism, we employ a Channel DeCorrelation Regularization (CDCR) to minimize the redundancy between the components of the compared vectors, which facilitate learning a hierarchical representation.
The proposed modules are orthogonal to the existing pretraining techniques~\cite{hero,clipbert,oscar,univl,frozen} and can be easily implemented by a few lines of code in modern libraries.

We validate the retrieval ability of our approach on multiple text-video retrieval datasets. 
Our experimental results outperform the state-of-the-art methods under widely used benchmarks, \textit{e.g.}, surpassing CLIP4Clip~\cite{clip4clip} largely by $+2.9\%$, $+3.1\%$, $+7.9\%$, $+2.3\%$, $+2.8\%$ and $+6.5\%$ R@1 on the MSR-VTT~\cite{msrvtt}, MSVD~\cite{msvd}, VATEX~\cite{vatex}, LSMDC~\cite{lsmdc}, AcitivityNet~\cite{anet}, and DiDeMo~\cite{didemo}, respectively.
Notably, using ViT-B/16~\cite{clip} and QB-Norm~\cite{qbnorm} post processing, our best model can achieve $53.3\%$ T2V R@1 and $56.2\%$ V2T R@1, surpassing all existing single-model entries.

Our empirical analysis suggests that there is vast room for improvement in the design of interaction mechanisms. 
The findings used in this paper make some initial headway in this direction. We hope that this study will spur further investigation into the operational mechanisms used in modeling interaction.

\section{Related Work}
\noindent\textbf{Feature Representation for Text-Video Retrieval.}
In recent studies~\cite{ce,clip,sclip,clip4clip}, the text and video inputs are usually considered separately for efficient deployment through a Bi-Encoder architecture~\cite{biencoder,colbert}.
The text encoder absorbs progress in the NLP field, and is upgraded from the early Word2Vec~\cite{w2v} to the BERT-like models~\cite{bert,roberta}.
While for the video input, due to its rich semantic content, researchers utilize multi-modality features~\cite{mmt} and a variety of pre-trained experts~\cite{ce} to boost performance.
Recently, CLIP~\cite{clip} proposes a concise contrastive learning method and trains on a large-scale dataset with 400 million pairs to obtain the generic cross-modal representation.
Indeed, in the span of just a few months, several CLIP-based TVR methods~\cite{clip4clip,clip2tv,clip2video} constantly refresh the state-of-the-art results on all benchmarks.
Our approach also benefits from existing pre-training, while with mainly focusing on the design of the interaction module. 
We conduct extensive and fair ablation studies to evaluate the effectiveness of our approach.

\noindent\textbf{Interaction Mechanism for Text-Video Retrieval.}
Over the past few years, most TVR studies have focused on improving performance by increasing the power of text and visual encoders~\cite{teachtext,mmt,hit,ce}. 
Thus the simple dot-product interaction is widely adopted for computing global similarity~\cite{clip4clip}. 
The time-series video features are aggregated with average pooling, LSTM~\cite{lstm}, or temporal transformer~\cite{clip4clip}.
Recent works~\cite{teachtext,mmt,ce} on the Mixture of Experts (MoE~\cite{moe}) show that integrating diverse domain experts can improve the overall model's capability with a voting procedure.
In~\cite{hit}, hierarchical feature alignment methods help models utilize information from different dimensions.
The pioneering work, JSFusion~\cite{jsfusion}, proposes to measure dense semantic similarity between all sequence data and learn a sophisticated attentions network. 
The Cross Transformer~\cite{hero,univl,clip4clip} establish the multi-modality correspondence by joint encoding texts and videos, which can capture both intra- and inter-modality context. 
We conduct the empirical study on the latest instantiation of interaction.
Our work is also related to several approaches that analyze the interaction mechanism on Information Retrieval (IR)~\cite{colbert} and Text-Image Retrieval~\cite{vilt}.
This work targets a deeper understanding of the interaction mechanism for Text-Video Retrieval in a new perspective.

\noindent\textbf{Contrastive Learning for Text-Video Retrieval.}
A common underlying theme that unites retrieval methods is that they aim to learn closer representations for the labelled text-video pairs.
The triplet loss based models\cite{support,ce} use a max-margin approach to separate positive from negative examples and perform intra-batch comparisons.
With the popularity of self-supervised learning (SSL)~\cite{ssl1,ssl2,infonce}, InfoNCE~\cite{infonce} have dominated recent retrieval tasks, which mainly maximizes the diagonal similarity and achieves superior performance for noisy data.
The Barlow Twins method~\cite{barlow} have recently been proposed as a new solution for SSL. 
They use a covariance matrix to indicate inter-channel redundancy.
We employ a sequential channel decorrelation regularization to reduce inter-channel redundancy and competition, which fits the assumptions of hierarchical semantics in Text-Video Retrieval.

\section{Methodology}
\label{sec:method}
\begin{figure}[t]
\centering
\includegraphics[width=0.99\linewidth]{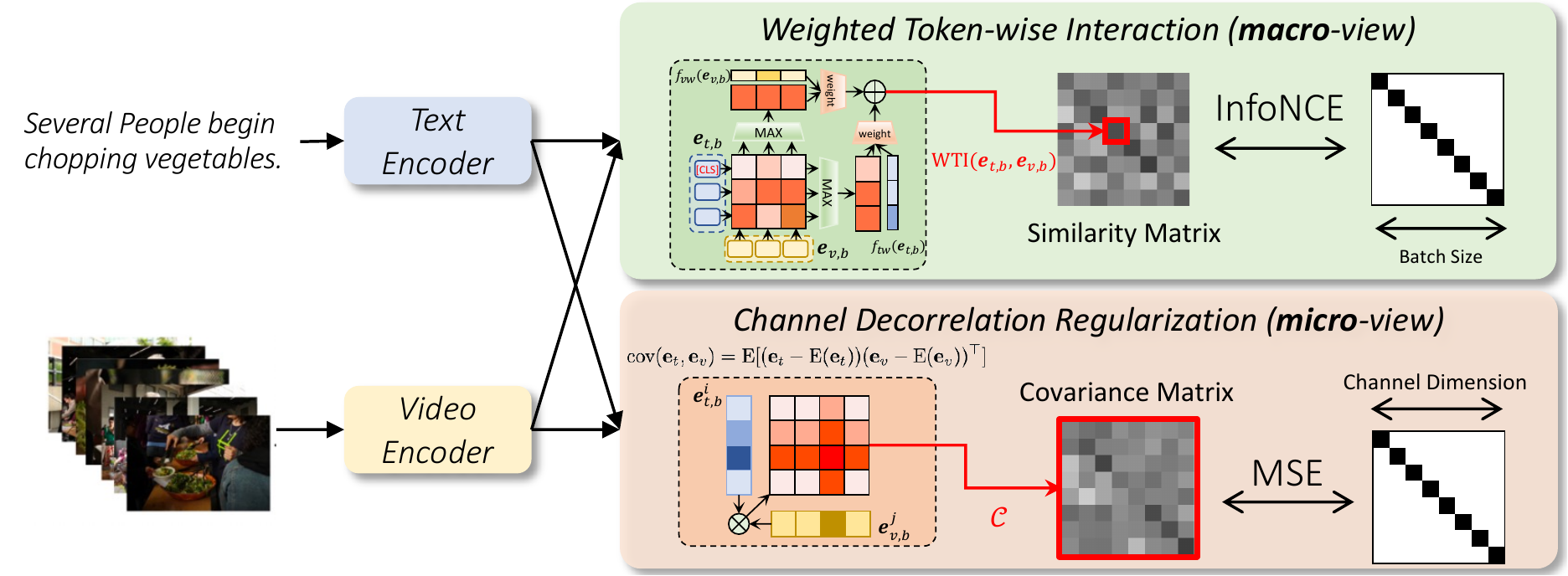}
\caption{Overview of the proposed disentangled representation learning design for Text-Video Retrieval. 
The pipeline of our method consists of two components: the Weighted Token-wise Interaction (WTI) block that exhaustively matching all sequential representation for the sentence and video; the Channel DeCorrelation Regularization (CDCR) minimizes the redundancy for the sequential representation.}
\label{fig:pipeline}
\end{figure}

The framework of the proposed Text-Video Retrieval (TVR) is shown in Figure~\ref{fig:pipeline}. 
Our method first generates sequential feature representation for both text and video. 
In the cross-modality interaction stage, the network will reason the dense correlation for all pair-wise features and dynamically adjust weights based on the token contents.
We additionally introduce a channel decorrelation regularization to minimize the redundancy of these features.
Consequently, given a text query $\mathbf{t}=\{t^{i}\}_{i=1}^{N_{t}}$ with $N_t$ tokens, and a set of video documents $\mathcal{V}=\{\mathbf{v}_n\}_{n=1}^{N}$, the text-video retrieval is formulated as a cross-modality similarity measurement, $\mathcal{S}(\mathbf{t}, \mathbf{v})$. 
We can extract text features $\mathbf{e}_t$ and video features $\mathbf{e}_v$ through text encoder $E_t(\mathbf{t}, \theta_t)$ and video encoder $E_v(\mathbf{v}, \theta_v)$, respectively.
Different from the global dot-product operation with single vector, our embedding features retain the sequential structure of text and video frames with $\mathbf{e}_t \in \mathbb{R}^{N_t\times D}$, $\mathbf{e}_v \in \mathbb{R}^{N_v\times D}$, where $N_v$ is the sampled video frames and $D$ is the feature dimension. 
We further introduce a Weighted Token-wise Interaction (WTI) module to estimate the cross-modality correlation and discover potential activate tokens.
In order to improve the domain generalization, we leverage the covariance matrix to indicate the redundancy of features and employ a simple Channel Decorrelation Regularization (CDCR) to improve the performance.
The decorrelation loss implicitly alleviates an overlooked channel competition, achieving impressive performance in retrieval more complex sentences.

\subsection{Feature Extractor}
We utilize the efficient Bi-Encoder architecture for feature extraction.
For each query sentence $\mathbf{t}$, we add the \texttt{[CLS]}  and \texttt{[SEP]} token to indicate the start and the end of the sentence, and adopt the pretrained BERT-BASE~\cite{bert} to encode the text representation $\mathbf{e}_t=E_{t}(\mathbf{t})$.
For each video $\mathbf{v}$, we uniformly select $N_v$ frames as keyframes and employ off-the-shelf transformer-based networks, \textit{e.g.} ViT~\cite{vit} , to extract sequential features $\mathbf{e}_v=E_{v}(\mathbf{v})$.

Retrieval across modalities benefits from large-scale pre-training tasks~\cite{colbert,bert,clip}.
In this paper, we mainly focus on the design of the interaction module rather than the pretrained network. 
Thus our feature extractors are initialized from the CLIP~\cite{clip}, and we finetune in an end-to-end manner.

\subsection{Study of Interaction Mechanisms }
\begin{table}[t]
\centering
\caption{Comparisons of different interaction mechanisms. $N$ denotes number of video documents ($N$ is large and depends on applications); $D$ denotes representation dimension ($D=512$, by default); $N_t$ and $N_v$ denote length of text token and frame token, $N_{v+t}$ is the sum of $N_t$ and $N_v$ ($N_t = 32, N_v = 12$); $L$ denotes number of network layer and $S$ denotes feature levels ($L=4, S=3$, by default).}
\begin{tabular}{l|c|c| c|c}
\toprule
Interaction & Contents    & Function          & Computational Complexity           & Memory   \\
\hline
DP   & single  & parameter-free       & $\mathcal{O}(ND)$              & $\mathcal{O}(ND)$  \\
HI                   & multi-level    & light-parameter & $\mathcal{O}(NSD)$             & $\mathcal{O}(NSD)$ \\
MLP                   & single    & black-box            & $\mathcal{O}(N(D^2L+D))$        & $\mathcal{O}(ND)$  \\
XTI      & token-wise   & black-box            & $\mathcal{O}(N(D^2 (N_{t+v}) + N_{t+v}^2D)L)$ & $\mathcal{O}(NN_{v}D)$ \\
TI                    & token-wise   & parameter-free       & $\mathcal{O}(NN_tN_vD)$            & $\mathcal{O}(NN_{v}D)$ \\
WTI                   & token-wise   & light-parameter & $\mathcal{O}(N(N_tN_vD+N_{t+v}))$            & $\mathcal{O}(NN_{v}(D+1))$\\
\bottomrule
\end{tabular}
\label{tab:cmp}
\end{table}%

To facilitate our study, we develop a generalized interaction formulation that is able to represent various module designs and show a comparison in Table~\ref{tab:cmp}.
We then show how the existing interaction mechanisms can be represented within this formulation, and how ablations can be conducted using this formulation with respect to different interaction module elements.

\noindent{\textbf{Single Vector Dot-product Interaction:}}
When measuring feature similarity, researchers use simple dot-product operation, which is an intuitive solution.
Specifically, two global representations are compared in the $\ell_2$ normalized embedding space:
\begin{equation}
\operatorname{DP}(\mathbf{e}_t, \mathbf{e}_v)= \frac{(\mathbf{e}_t^{\text{[CLS]}})^{\top}\cdot   \bar{\mathbf{e}}_v}{\|\mathbf{e}_s^{\text{[CLS]}}\|_{2} \cdot  \| \bar{\mathbf{e}}_v \|_{2}},
\end{equation}
where $\mathbf{e}_t^{\text{[CLS]}}$ is the text feature for \texttt{[CLS]} token and $\bar{\mathbf{e}}_v = \frac{1}{N_v} \sum_{i=1}^{N_V} \mathbf{e}_v^{i}$ is the average video representation. 
The dot-product method is widely used in biometrics~\cite{arcface,reid}, text-image retrieval~\cite{flickr30k}, and text document retrieval~\cite{dssm}.
This interaction method is extremely efficient and can be accelerated by the ANN library, \textit{e.g.}, FAISS~\cite{faiss}.
While, due to the single-vector representation, the sequential structure is heavily coupled in encoder. 
Compressing the long sequence into a single vector will push the network to learn extremely complex abstractions and may easily miss out fine-grained clues. 
The dot-product function also encourages smooth latent spaces, which will discourage the hierarchical semantics learning.

\noindent{\textbf{Hierarchical Interaction:}}
Similar to the global feature representation, multi-layer~\cite{hit}, multi-modal~\cite{mmt} and multi-experts~\cite{ce} use the gated fusion mechanism to ensemble the hierarchical representation:
\begin{equation}
\operatorname{HI}(\mathbf{e}_t, \mathbf{e}_v)=\sum_{s=1}^{S} w (\mathbf{e}_{t,s}^{\text{[CLS]},s})^\top \bar{\mathbf{e}}_{v,s},
\end{equation}
where $w_s$ can be a normalized weight or a binarized gated unit.
We note that this hierarchical interaction actually contains two essential factors: one is the retrieval-oriented feature pool, and the other is the dynamic fusion block.
The content of the comparison has been greatly enriched from expert models. While  the learning of the experts requires labor intensive manual annotation.

\noindent{\textbf{MLP on Global Vector:}}
In order to improve the nonlinear measurement ability of the interaction block, researchers~\cite{spm,clip4clip} propose to use neural network to learn the metric, among which the representative work~\cite{spm} builds a Multilayer Perceptron (MLP) measurement.
They directly concatenate the compared features as a whole, and optimize the similarity through MLP:
\begin{equation}
\operatorname{MLP}(\mathbf{e}_t, \mathbf{e}_v)=f_\theta([\mathbf{e}_t^{\text{[CLS]}},\bar{\mathbf{e}}_v]).
\end{equation}
where $[,]$ denotes concatenation operation. 
Although the neural network brings a nonlinear metric space, it also operates in a black-box setting. 
The pure-parameterizative method needs to consume massive amount of labelled data in the training process. 
And when deployed on large-scale (billions) documents, even two fully connected layers will bring prohibitively computational overhead.

\noindent{\textbf{Cross Transformer Interaction:}}
The standard self-attention~\cite{attention} is adopted for cross-modality matching, which can handle variable-length inputs:
\begin{equation}
\operatorname{XTI}(\mathbf{e}_t, \mathbf{e}_v)=\text{MHA}_\theta([\mathbf{e}_t,\mathbf{e}_v]).
\end{equation}

The Cross Transformer aims to capture and model the inter-modal relations between texts and videos by exchanging key-value pairs in the multi-headed attention (MHA) mechanism.
In Section~\ref{sec:ablation}, we observe that the cross transformer interaction is difficult to optimize and usually occurs performance degradation on video retrieval datasets.
Furthermore, compared with the dot-product operation, multi-head attention brings thousand times of the computational overhead, as shown in Table~\ref{tab:cmp} and Table~\ref{tab:abl_interaction_loss_msrvtt}.

\noindent{\textbf{Token-wise Interaction:}}
Recently, ColBERT~\cite{colbert} and FILIP~\cite{filip} propose token-wise interaction for document and image retrieval. We introduce this token-wise interaction to TVR, which cleverly solves the local context matching problem (Assumption (1) in Figure~\ref{fig:intro}):
\begin{equation}
\operatorname{TI}(\mathbf{e}_t, \mathbf{e}_v)=\left(\sum_{i=1}^{N_t} \max_{j=1}^{N_v}(\tilde{\mathbf{e}}_t^{i})^{\top}\tilde{\mathbf{e}}_v^{j} + \sum_{j=1}^{N_v} \max_{i=1}^{N_t}(\tilde{\mathbf{e}}_t^{i})^{\top}\tilde{\mathbf{e}}_v^{j} \right)/2,
\label{eq:ti}
\end{equation}
where $\tilde{\mathbf{e}} = \mathbf{e}^i /  \| \mathbf{e}^i \|_{2}$ is the channel-wise normalization operation. The token-wise interaction is a parameter-free operation and allows efficient storing and indexing~\cite{colbert}.

\noindent{\textbf{Weighted Token-wise Interaction:}}
Intuitively, not all words and video frames contribute equally. 
We provide an adaptive approach to adjust the weight magnitude for each token:
\begin{equation}
\operatorname{WTI}(\mathbf{e}_t, \mathbf{e}_v)=\left(\sum_{i=1}^{N_t} f_{tw,\theta}^{i}(\mathbf{e}_t) \max_{j=1}^{N_v}(\tilde{\mathbf{e}}_t^{i})^{\top}\mathbf{e}_v^{j} + \sum_{j=1}^{N_v} f_{vw,\theta}^{j}(\mathbf{e}_v)\max_{i=1}^{N_t}(\tilde{\mathbf{e}}_t^{i})^{\top}\tilde{\mathbf{e}}_v^{j}  \right)/2,
\label{eq:wti}
\end{equation}
where $f_{tw,\theta}$ and $f_{vw,\theta}$ are composed of classic MLP and a SoftMax function.
The adaptive block is lightweight and takes a \textbf{single}-modality input, allowing offline pre-computation for large-scale video documents. 
In the online process, the attention module introduces negligible computational overhead, as shown in Table~\ref{tab:abl_interaction_loss_msrvtt}.
Our method inherits the general matching priors and is empirically verified to be more effective and efficient.

WTI can be easily implemented by a few lines of code in modern libraries.
Algorithm~\ref{alg:pytorch} shows a simplified code based on PyTorch~\cite{pytorch}.

\begin{algorithm}[tb]
   \caption{PyTorch-style pseudocode for Weighted Token-wise Interaction.}
   \label{alg:pytorch}
   
    \definecolor{codeblue}{rgb}{0.25,0.5,0.5}
    \lstset{
      basicstyle=\fontsize{7.2pt}{7.2pt}\ttfamily\bfseries,
      commentstyle=\fontsize{7.2pt}{7.2pt}\color{codeblue},
      keywordstyle=\fontsize{7.2pt}{7.2pt},
    }
\begin{lstlisting}[language=python]
# t: text input                 v: video input
# f_t: text encoder network     f_v: video encoder network
# f_tw: text weight network     f_vw: video weight network
# B: batch size                 D: dimensionality of the embeddings
# N_t: text token size          N_v: frame token size

def weighted_token_wise_interaction(t, v):    
    # compute embeddings
    e_t = f_t(t) # BxN_txD
    e_v = f_v(v) # BxN_vxD
    
    # generate fusion weights
    text_weight = torch.softmax(f_tw(e_t), dim=-1)    # BxN_t
    video_weight = torch.softmax(f_tv(e_v), dim=-1)    # BxN_v
    
    # normalize representation
    e_t = e_t / e_t.norm(dim=-1, keepdim=True)  # BxN_txD
    e_v = e_v / e_v.norm(dim=-1, keepdim=True)  # BxN_vxD
    
    # token interaction
    logits = torch.einsum("atc,bvc->abtv", [e_t, e_v])  # BxBxN_txN_v
    t2v_logits = logits.max(dim=-1)[0]  # BxBxN_txN_v -> BxBxN_t
    t2v_logits = torch.einsum("abt,at->ab", [t2v_logits, text_weight])  # BxBxN_t -> BxB
    v2t_logits = logits.max(dim=-2)[0]  # BxBxN_v
    v2t_logits = torch.einsum("abv,bv->ab", [v2t_logits, video_weight])  # BxBxN_v-> BxB
    
    retrieval_logits = (t2v_logits + v2t_logits) / 2.0  # BxB
    
    return retrieval_logits
\end{lstlisting}
\end{algorithm}

\subsection{Channel Decorrelation Regularization}
Given a batch of $B$ video-text pairs, WTI generates an $B\times B$ similarity matrix.
The Text-Video Retrieval is trained in a supervised way. 
We employ the InfoNCE loss~\cite{infonce} to maximize the similarity between labelled video-text pairs and minimize the similarity for other pairs:
\begin{equation}
\begin{gathered}
\mathcal{L}_{\text {InfoNCE}} = \mathcal{L}_{v 2 t} + \mathcal{L}_{t 2 v}\\
\mathcal{L}_{v 2 t} =-\frac{1}{B} \sum_{i}^{B} \log \frac{\exp \left(\operatorname{WTI}\left(\mathbf{e}_{t,i}, \mathbf{e}_{v,i}\right)/\tau\right)}{\sum_{j}^{B} \exp \left(\operatorname{WTI}\left(\mathbf{e}_{t,i}, \mathbf{e}_{v,j}\right)/\tau\right)} \\
\mathcal{L}_{t 2 v} =-\frac{1}{B} \sum_{i}^{B} \log \frac{\exp \left(\operatorname{WTI}\left(\mathbf{e}_{t,i}, \mathbf{e}_{v,i}\right)/\tau\right)}{\sum_{j}^{B} \exp \left(\operatorname{WTI}\left(\mathbf{e}_{t,j}, \mathbf{e}_{v,i}\right)/\tau\right)},
\end{gathered}
\end{equation}
where $\tau$ is temperature hyper-parameter.

The contrastive loss provides a \textit{macro} objective to optimize the global similarity.
Referring expression comprehension, the multi-modality retrieval often requires semantic information from \textit{micro}-views, \textit{e.g.}, the channel-level.
Inspired from self-supervised learning methods~\cite{barlow}, we utilize the covariance matrix to measure the redundancy between features and employ a simple $\ell_2$-norm minimization to optimize the hierarchical representation:
\begin{equation}
\begin{gathered}
\mathcal{L}_{\text{CDCR}} =\sum_{i}\left(1-\mathcal{C}^{i i}\right)^{2}+\alpha  \sum_{i} \sum_{j \neq i} (\mathcal{C}^{i j})^{2}\\
\mathcal{C}^{i j} \triangleq \frac{\sum_{b} \mathbf{e}_{t, b}^{(i)} \mathbf{e}_{v, b}^{(j)}}{\sqrt{\sum_{b}\left(\mathbf{e}_{t, b}^{(i)}\right)^{2}} \sqrt{\sum_{b}\left(\mathbf{e}_{v, b}^{(j)}\right)^{2}}},
\end{gathered}
\end{equation}
where $\mathbf{e}_{t, b}^{(i)}$ is the $i$-th channel of $b$-th text feature $\mathbf{e}_{t, b}$. The coefficient $\alpha$ controls the magnitude of the redundancy term.
The total training loss $\mathcal{L}_{\text{all}}$  is  defined as: 
\begin{equation}
\begin{gathered}
\mathcal{L}_{\text{all}} = \mathcal{L}_{\text{InfoNCE}} + \lambda \mathcal{L}_{\text{CDCR}},
\end{gathered}
\end{equation}
where $\lambda$ is the weighting parameter.
Surprisingly, empirical results show that the cross-modality channel decorrelation regularization brings significant improvements for all interaction mechanisms, as shown in Table~\ref{tab:abl_interaction_loss_msrvtt}.

\section{Experiments}
\label{sec:experiments}
In this section, we firstly present the configuration details of our algorithm and then conduct thorough ablation experiments to explore the relative importance of each component in our method on MSR-VTT~\cite{msrvtt}.
Finally we conduct comprehensive experiments on six benchmarks: MSR-VTT~\cite{msrvtt}, MSVD~\cite{msvd}, VATEX~\cite{vatex},  LSMDC~\cite{lsmdc}, ActivityNet~\cite{anet} and DiDeMo~\cite{didemo}.  

\subsection{Experimental Settings} 
\textbf{Dataset}: We conduct experiments on six benchmarks for video-text retrieval tasks including:
\begin{itemize}
\item \textbf{MSR-VTT}~\cite{msrvtt} contains 10,000 videos with 20 captions for each. We report results on 1k-A which adopts 9,000 videos with all corresponding captions for training and utilizes 1,000 video-text pairs as test.
\item \textbf{MSVD}~\cite{msvd} includes 1,970 videos with 80,000 captions.  We report results on split set,  where train, validation and test are 1200, 100 and 670 videos.
\item\textbf{VATEX}~\cite{vatex} contains 34,991 videos with multilingual annotations. The training split contains 25,991 videos.  We report the results on the split set includes 1500 videos for validation and 1500 videos for test. 
 \item\textbf{LSMDC}~\cite{lsmdc} contains 118081 videos and equal captions extracted from 202 movies with a split of 109673,  7408, and 1000 as the train, validation, and test set.  Every video is selected from movies ranging from 2 to 30 seconds.
 \item\textbf{ActivityNet}~\cite{anet} consists of 20,000 YouTube videos. We concatenate all descriptions of a video to a single query and evaluate on the `val1' split. 
 \item\textbf{DiDeMo}~\cite{didemo} contains 10,000 videos annotated with 40,000 sentences. All sentence descriptions for a video are also concatenated into a single query for text-video retrieval.
\end{itemize}

\noindent\textbf{Evaluation Metric:} We follow the standard retrieval task~\cite{clip4clip} and adopt Recall at rank K (R@K),  median rank (MdR) and mean rank (MnR) as metrics.  Higher R@K and lower MdR or MnR indicates better performance.

\noindent\textbf{Implementation Details}: We utilize the standard Bi-Encoder from CLIP~\cite{clip} as pre-trained feature extractor. 
Concretely, the vision encoder is comprised of a vanilla ViT-B/32~\cite{vit} and 4-layers of temporal transformer blocks. Each block employs 8 heads and 512 hidden channels. The temporal position embedding and network weight parameters are initialized from the CLIP's text encoder.
The fixed video length and caption lengths are 12 and 32 for MSR-VTT, MSVD, VATEX, LSMDC and 64 and 64 for ActivityNet and DiDeMo.
Following~\cite{clip}, the special tokens, \texttt{[CLS]} and \texttt{[SEP]}, and the text tokens are concatenated as inputs to a 12-layer linguistic transformer~\cite{bert,clip}.
We leverage the MLP to capture the weight factor for tokens, and our ablation studies suggest a 2-layer structure.
The adaptive module is initialized from scratch with random weights.

\noindent\textbf{Training Schedule}:  For fair comparisons with our baseline,  we follow training schedules from CLIP4Clip~\cite{clip4clip}.
The network is optimized by Adam~\cite{adam} with a batch size of 128 in 5 epochs. 
The initial learning rate for vision encoder and text encoder are set $1\times 10^{-7}$, and the initial learning rate for the temporal transformer and the adaptive module are set to $1\times10^{-4}$. 
All learning rates follow the cosine learning rate schedule with a linear warmup. 
We apply a weight decay regularization to all parameters except for bias, layer normalization, token embedding, positional embedding and temperature.
During training, we set the temperature $\tau=100$, CDCR weight $\alpha=0.06$ and overall weighting parameter $\lambda=0.001$.   

\subsection{Ablation Study}
\label{sec:ablation}

\begin{table}[tbp]
\centering
\caption{Ablation of Disentangled Representation on the 1K validation set of MSR-VTT~\cite{msrvtt}. Time shows inference speed for indexing 1 million video documents on a Tesla V100 GPU.}
\vspace*{-2mm}
\setlength{\tabcolsep}{2.pt}
\begin{tabular}{lccccccccc}
\toprule
\multirow{2}{*}{Interaction} & \multicolumn{4}{c}{InfoNCE~\cite{infonce}}& \multicolumn{4}{c}{+\textbf{CDCL}} &  \multirow{2}{*}{\begin{tabular}[c]{@{}l@{}}Time\\ (ms)\end{tabular}}
\\
& R@1$\uparrow$ & R@5$\uparrow$& R@10$\uparrow$ & MnR$\downarrow$ &  R@1$\uparrow$ & R@5$\uparrow$ & R@10$\uparrow$ & MnR$\downarrow$ &\\
\hline
DP&  42.8&72.1 &81.4& 16.3& 44.2  & 72.7 & 82.0 & 14.5& \textbf{415}\\
HI &   43.5& 72.9& 81.7& 16.1& 44.1 &  72.6&82.8 & 14.1& 531\\
MLP &  29.3& 54.8& 64.2& 33.8 & -& -& - &-& 25,304\\
XTI &  41.8 &71.2&82.7& 16.2& - &- &- &-& 80,453\\
\rowcolor{Gray}
\textbf{TI} &44.8&\textbf{73.7}& 82.9 &13.5 &45.5 &72.0& 82.5  & 13.3& 536\\
\rowcolor{Gray}
\textbf{WTI} &  \textbf{46.3} &\textbf{73.7}& \textbf{83.2} &\textbf{13.0} &  \textbf{47.4} &  \textbf{74.6} & \textbf{83.8} & \textbf{12.8}& 565 \\ \hline
DP$_{\text{+ViT-B/16}}$& 45.9  & 73.8 & 82.3 & 13.8 & 46.6  & 73.3 & 82.8  & 13.4 & \textbf{415}\\
\rowcolor{Gray}
\textbf{TI}$_{\text{+ViT-B/16}}$&47.3&\textbf{76.7}& \textbf{84.8}&13.7&49.1&75.7& 85.1  &12.7 &536\\
\rowcolor{Gray}
\textbf{WTI}$_{\text{+ViT-B/16}}$&  \textbf{48.8}  & 76.1 & 84.3 & \textbf{13.5}&  \textbf{50.2} & \textbf{76.5} & \textbf{84.7} & \textbf{12.4} & 565\\
\bottomrule
\end{tabular}
\label{tab:abl_interaction_loss_msrvtt}
\vspace{-2mm}
\end{table}

We quantitatively evaluate the key components, Weighted Token-wise Interaction (WTI) and Channel DeCorrelation Regularization (CDCR), on MSR-VTT 1K~\cite{msrvtt}.
Table~\ref{tab:abl_interaction_loss_msrvtt} summarizes the ablation results.

\noindent\textbf{Effect of Weighted Token-wise Interaction:} 
Compared with the Single-Vector Dot-Product Interaction (DP), the Token-wise Interaction (TI) only adds a dense correlation, while the performance boosts $+2.0\%$ R@1. 
Our Weighted Token-wise Interaction (WTI) dramatically improves the performance by an R@1 of $3.5\%$, which shows that the proposed interaction structure indeed helps the model to adequately exploit pair-wise correlations.
Compared with other light-parameter function, our method improves by $+2.8\%$ R@1 over Hierarchical Interaction (HI).
An explanation is that the token representation is more valuable than the multi-level features for the single-scale ViT~\cite{vit}.
As observed in~\cite{clip4clip}, the MLP interaction (MLP) and Cross Transformer Interaction (XTI), under black-box function, would lead to degradation problems.
Further, by adopting ViT-B/16~\cite{clip}, WTI greatly promotes the T2V R@1 to $48.8\%$.

\noindent\textbf{Effect of Channel DeCorrelation Regularization:} 
In Table~\ref{tab:abl_interaction_loss_msrvtt}, we evaluate the importance of Channel Decorrelation Regularization on different interaction structures except block-box function.
Surprisingly, CDCR increases performance significantly for all interactions, \textit{i.e.} by $+1.4\%$, $+0.6\%$, $+0.7\%$ and $+1.1\%$ for DP, HI, TI and WTI respectively.
The micro channel-level regularization makes it easier to leverage semantic hierarchy.
\begin{table}[t]
\centering
\caption{Effect of dual-path and layers for weight model on MSR-VTT 1K~\cite{msrvtt}.  }
\vspace*{-2mm}
\setlength{\tabcolsep}{0.7pt}
\begin{tabular}{lcccccccccccc}
\toprule
 \multirow{2}{*}{Dual}
  & \multicolumn{4}{c}{t2v}& \multicolumn{4}{c}{v2t}& \multicolumn{4}{c}{\textbf{t2v+v2t}}\\
 \cmidrule(r){2-5}
 \cmidrule(r){6-9}
  \cmidrule(r){10-13}
 & R@1$\uparrow$ & R@5$\uparrow$& R@10$\uparrow$ & MnR$\downarrow$ &  R@1$\uparrow$ & R@5$\uparrow$ & R@10$\uparrow$ & MnR$\downarrow$ &  R@1$\uparrow$ & R@5$\uparrow$ & R@10$\uparrow$ & MnR$\downarrow$\\   
\hline
\textbf{TI}  & 43.4  &  71.0  & 80.2  & 16.1 & 43.5  &  71.8  & 82.7  & 14.0 & \textbf{44.8}  &  \textbf{73.7}  & \textbf{82.9}  & \textbf{13.5}\\
\textbf{WTI} & 45.4  & 72.3  & 81.7  & 13.4 & 45.3  & \textbf{74.6}  & \textbf{83.3}  & 13.6 & \textbf{46.3}  &  73.7  & 83.2  & \textbf{13.0}\\
\bottomrule
Layers & \multicolumn{4}{c}{1FC}& \multicolumn{4}{c}{\textbf{2FC}}& \multicolumn{4}{c}{3FC}\\ 
\hline
\textbf{WTI} & \textbf{46.3}  & \textbf{73.9}  & 82.9  & 13.7 & \textbf{46.3} & 73.7  & \textbf{83.2}  & \textbf{13.0} & 45.7  &  72.9  & 81.4  & 14.0\\
\bottomrule
\end{tabular}
\label{tab:abl_dual_msrvtt}
\vspace{-2mm}
\end{table}

\noindent\textbf{Effect of Dual-path Token-wise Retrieval:} 
In Table~\ref{tab:abl_dual_msrvtt}, we test the importance of dual-path interaction of t2v and v2t, subset of Eq.~\ref{eq:ti}. 
By adopting dual-path to compute the logits, our model provides stable gains by $+1.4\%$ and $+1.0\%$ for TI and WTI, respectively.

\noindent\textbf{Effect of MLP layers:}  By default, we simply set the weighted function with MLP, which is composed of FC+ReLU blocks.
In Table~\ref{tab:abl_dual_msrvtt}, we observe that R@1 is slightly better when applying 2FC layers and drop by 3FC, likely because a heavier structure suffers from over-fitting.

\begin{figure}[t]
\centering
\includegraphics[width=0.48\linewidth]{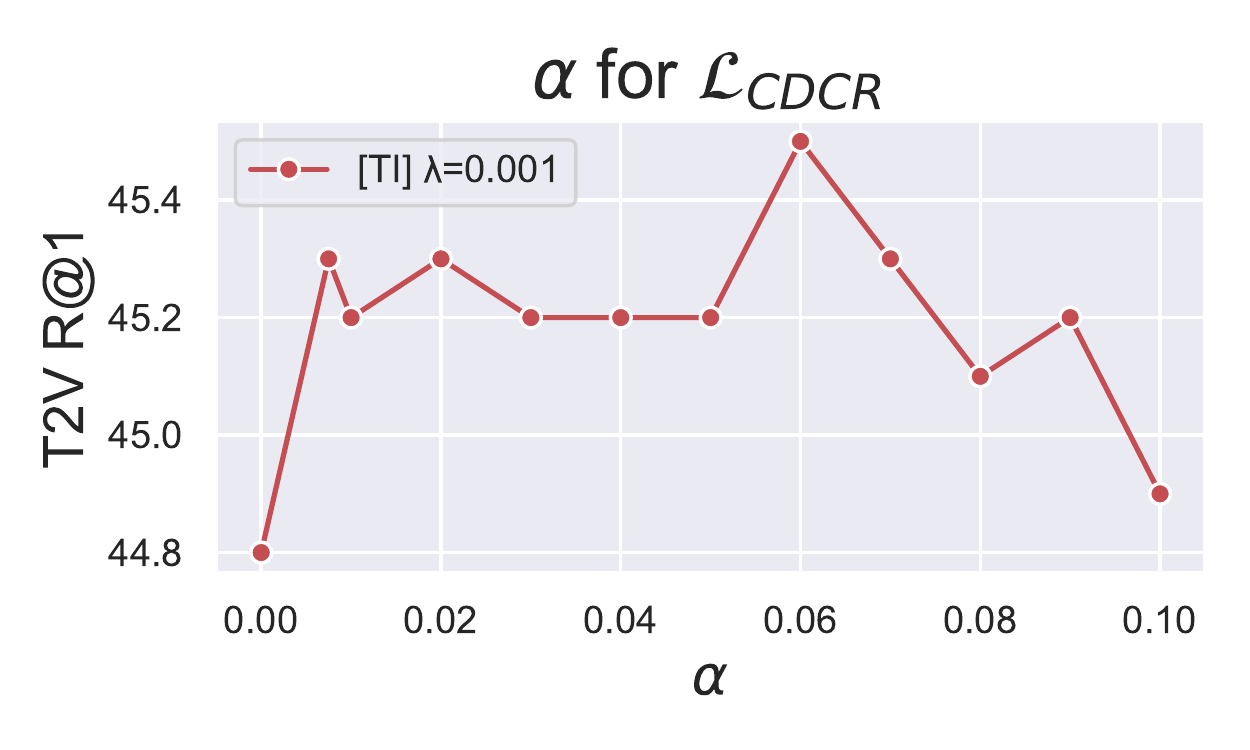}
\includegraphics[width=0.48\linewidth]{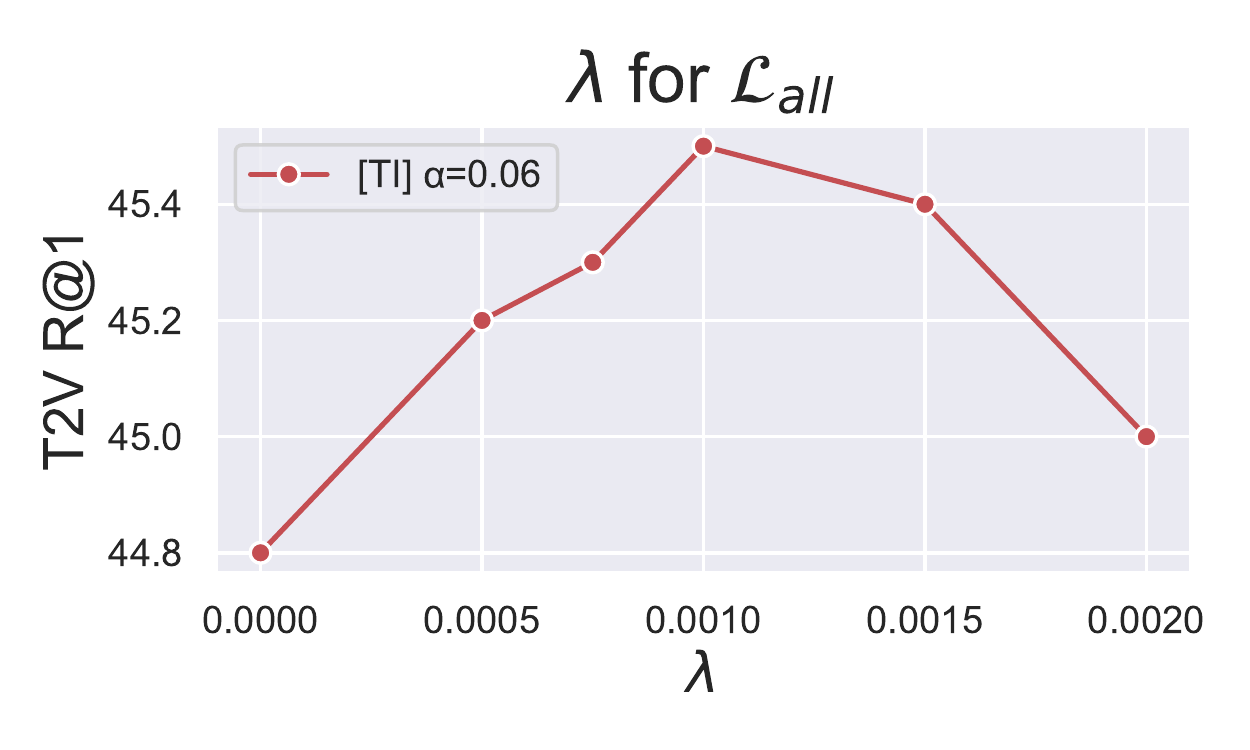}
\vspace{-4mm}
\caption{Hyper-parameter for feature decorrelation. (a) $\alpha$ for $\mathcal{L}_{\text{CDCR}}$ (b) $\lambda$ for $\mathcal{L}_{\text{all}}$.}
\vspace{-5mm}
\label{fig:hp}
\end{figure}
\begin{figure}[t]
\centering
\includegraphics[width=0.48\linewidth]{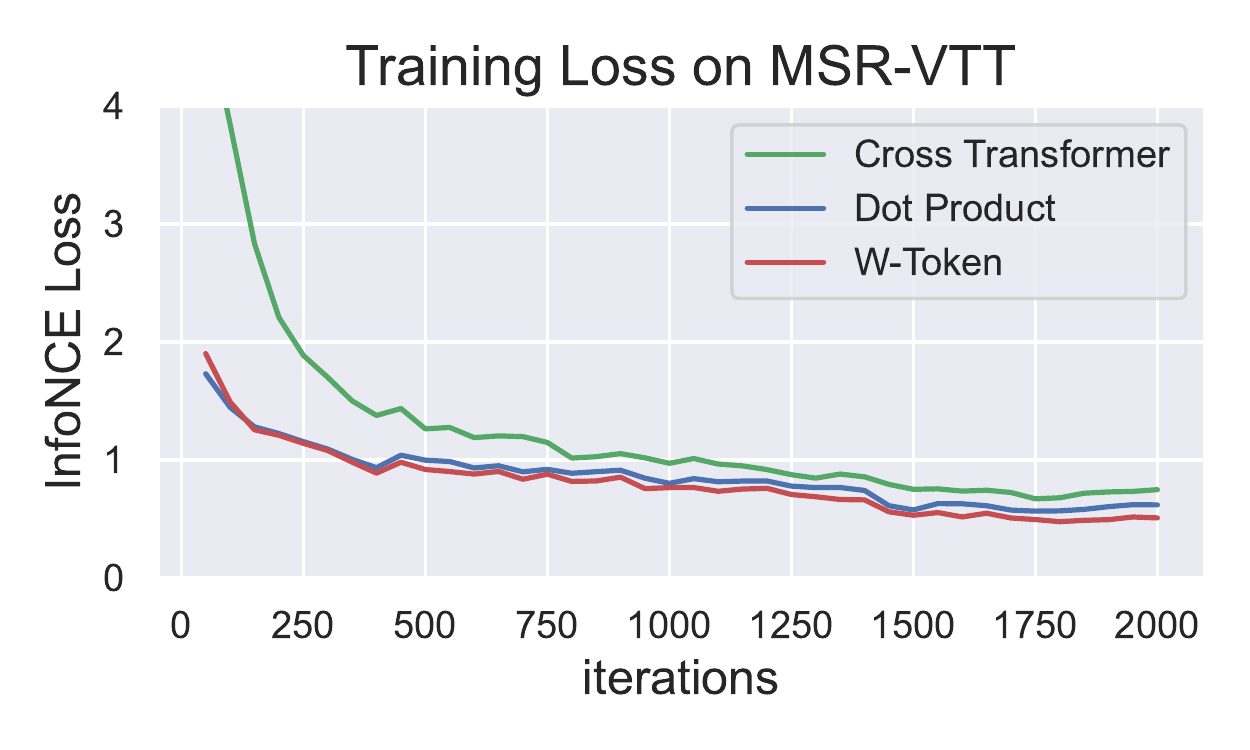}
\includegraphics[width=0.48\linewidth]{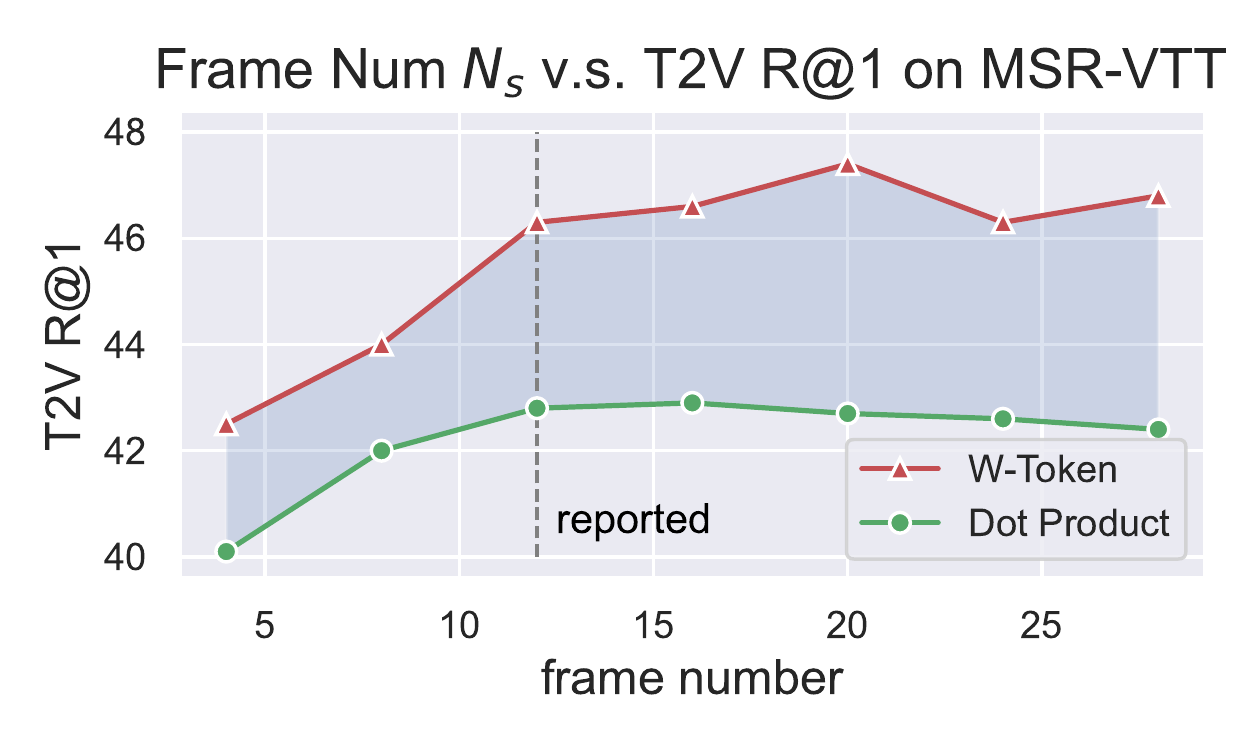}
\vspace{-4mm}
\caption{The training loss comparisons (a) and the influence of frame number (b).}
\label{fig:loss}
\end{figure}

\noindent\textbf{Hyper-parameter Selection for DeCorrelation:}  The parameter $\alpha$ and $\lambda$  specifies the scale and importance of the Channel DeCorrelation Loss.
We evaluate the scale range setting $\alpha \in [0,0.1]$ and $\lambda \in [0,0.002]$ as shown in Figure~\ref{fig:hp}. 
We find that R@1 is improved from $44.8\%$ to $45.5\%$ when $\alpha=0.06$ and saturated with $\alpha=0.07$.
We can get the best R@1 when the $\lambda$ is around 0.001.
As a result, we adopt  $\alpha = 0.06$ and $\lambda= 0.001$ in our model to achieve the best performance.

\noindent\textbf{Training Convergence Analysis:}  
To further reveal the source of the performance improvement, we present the training loss in Figure~\ref{fig:loss}. 
The training of pure-parameterizative interaction (Cross Transformer) is not trivial since it lacks the basic inductive biases for distance measurement. 
Inspired by this observation, we seek a light-weight way to leverage the heuristic property for better convergence, and therefore achieve a faster convergence rate.

\noindent\textbf{Benefit for More Frames:} 
Figure~\ref{fig:loss}(b) shows performance for DP and WTI with different frames. 
Compared with the naïve baseline DP, our WTI can achieve more significant gains, $+4.9\%$ v.s. $+2.8\%$, with more frames. For fair comparisons with CLIP4Clip~\cite{clip4clip}, we mainly report results at 12 frames.

\noindent\textbf{Inference Speed:}  Besides recall metric, we report the inference speed of all the interaction structure in Table~\ref{tab:abl_interaction_loss_msrvtt}.  Our lightweight interaction introduces negligible computational overhead, bringing remarkable improvements. Our approach reduce the inference time by several orders of magnitude, compared with the heavy cross transformer.

\subsection{Comparison with State-of-the-Arts }
\label{sec:sota}
\begin{table}[tbp]
\centering
\caption{Retrieval results on the validation set of MSR-VTT 1K~\cite{msrvtt}.
}
\vspace*{-2mm}
\small
\setlength{\tabcolsep}{2.pt}
\begin{tabular}{lcccccccc}
\toprule
 \multirow{2}{*}{Method} & \multicolumn{4}{c}{Text$\rightarrow$ Video}& \multicolumn{4}{c}{Video $\rightarrow$Text}\\
 \cmidrule(r){2-5}
 \cmidrule(r){6-9}
 & R@1$\uparrow$ & R@5$\uparrow$& R@10$\uparrow$ & MdR$\downarrow$ &  R@1$\uparrow$ & R@5$\uparrow$ & R@10$\uparrow$ & MdR$\downarrow$ \\
\hline
HERO~\cite{hero} &   16.8& 43.4& 57.7& - &- &- &-& - \\
UniVL~\cite{univl} &  21.2&49.6 &63.1& 6.0& - &- &- &- \\
ClipBERT~\cite{clipbert}&  22.0 &46.8 &59.9& 6.0& - &- &- &- \\
MDMMT~\cite{mmt} &   26.6& 57.1& 69.6& 4.0& 27.0& 57.5& 69.7 &3.7 \\
SUPPORT~\cite{support} &27.4& 56.3& 67.7 &3.0 &26.6 &55.1& 67.5 &3.0 \\
FROZEN~\cite{frozen}&  31.0& 59.5& 70.5& 3.0&  -&  - & -& -  \\
CLIP4Clip~\cite{clip4clip}&  44.5 &71.4& 81.6& \textbf{2.0} & 42.7& 70.9& 80.6& \textbf{2.0} \\
\hline
\rowcolor{Gray}
\textbf{Ours} & \textbf{47.4}  & \textbf{74.6}&  \textbf{83.8} & \textbf{2.0} & \textbf{45.3} & \textbf{73.9} & \textbf{83.3} & \textbf{2.0} \\
\rowcolor{Gray}
+ViT-B/16 & 50.2 & 76.5  & 84.7 &  1.0  & 48.9  & 76.3 & 85.4 & 2.0\\
\rowcolor{Gray}
+QB-Norm~\cite{qbnorm} &  53.3 & 80.3&  87.6& 1.0 & 56.2 & 79.9  & 87.4 & 1.0  \\
\bottomrule
\end{tabular}
\label{tab:sota_msrvtt}
\end{table}

In this subsection, we compare the proposed model with recent state-of-the-art methods on the six benchmarks, MSR-VTT~\cite{msrvtt}, MSVD~\cite{msvd}, VATEX~\cite{vatex}, LSMDC~\cite{lsmdc}, ActivityNet~\cite{anet} and DiDeMo~\cite{didemo}.

For \textbf{MSR-VTT} in Table~\ref{tab:sota_msrvtt}, our model significantly surpasses the ClipBERT~\cite{clipbert} by absolute $25.5\%$ R@1, reaching $47.4\%$ R@1 on MSR-VTT, indicating the benefits and necessity of large scale image-text pre-training for video-text retrieval.
We achieve $2.9\%$ improvement compared to CLIP4Clip~\cite{clip4clip}, which shows the benefit from token-wise interaction.
Our WTI, employing ViT-B/16 and QB-Norm~\cite{qbnorm}, yields a remarkable T2V R@1 $53.3\%$.

Table~\ref{tab:sota_sum} shows results for other benchmarks.
For \textbf{MSVD}, our model also demonstrates competitive performance with top-performing CLIP based model~\cite{clip4clip}. 
For \textbf{VATEX}, our approach achieves $+7.6\%$ R@1 improvement for text-video retrieval and $+3.8\%$ improvement for video-text retrieval.
For \textbf{LSMDC}, 
our approach achieves $+2.3\%$ R@1 improvements for text-video retrieval.

For \textbf{ActivityNet}, we outperform the state-of-the-art method by a large margin of $+2.8\%$ R@1 on text-video retrieval.   
For \textbf{DiDeMo}, we achieve remarkable performance $47.9\%$ R@1 and a relative performance improvement of $16.5\%$ in T2V compared to CLIP4Clip.   
It is worth noting that long sentences are used in ActivityNet and DiDeMo. The significant improvements on both datasets further proves the advantages of our disentangled representation.

Overall, the consistent improvements across different benchmarks strongly demonstrate the effectiveness of our algorithm. We hope that our studies will spur further investigation in modeling interaction.

\section{Conclusion}
\label{sec:conclusion}
In this work, we present an empirical study to give a better general understanding of interaction mechanisms for Text-Video Retrieval.
Then we propose a novel 
\begin{table}[H]
\centering
\caption{Retrieval results on the validation set of MSVD~\cite{msvd}, VATEX~\cite{vatex}, LSMDC~\cite{lsmdc}, ActivityNet~\cite{anet} and DiDeMo~\cite{didemo}.
}
\vspace*{-2mm}
\small
 \def\arraystretch{0.97}
\setlength{\tabcolsep}{3pt}
\begin{tabular}{lcccccccc}
\toprule
 & \multicolumn{4}{c}{Text$\rightarrow$ Video}& \multicolumn{4}{c}{Video $\rightarrow$Text}\\
 \cmidrule(r){2-5}
 \cmidrule(r){6-9}
Method & R@1$\uparrow$ & R@5$\uparrow$& R@10$\uparrow$ & MdR$\downarrow$ &  R@1$\uparrow$ & R@5$\uparrow$ & R@10$\uparrow$ & MdR$\downarrow$ \\
\hline
\multicolumn{9}{c}{Retrieval performance on MSVD~\cite{msvd}} \\
\hline
CE~\cite{ce}&  19.8 &49.0 &63.8 &6.0& -& - & - &-\\
SUPPORT~\cite{support} &   28.4& 60.0& 72.9& 4.0& -& -& -& - \\
FROZEN\cite{frozen}&  33.7 &64.7& 76.3& 3.0& -& -& -& -  \\
CLIP~\cite{clip}&  37.0 & 64.1 & 73.8 & 3.0 & 54.9  & 82.9  & 89.6  &\textbf{1.0}\\
CLIP4Clip\cite{clip4clip}&  45.2  &75.5 & 84.3  &2.0 &  48.4 & 70.3 & 77.2 & 2.0 \\
\rowcolor{Gray}
\textbf{Ours}&   \textbf{48.3} & \textbf{79.1 }& \textbf{87.3}& \textbf{2.0} & \textbf{62.3} & \textbf{86.3} & \textbf{92.2 } & \textbf{1.0}  \\
\rowcolor{Gray}
\textbf{Ours}$_{\text{+ViT-B/16}}$&   50.0  & 81.5 & 89.5 & 2.0 & 68.7   &92.5 & 95.6& 1.0   \\
\bottomrule
\multicolumn{9}{c}{Retrieval performance on VATEX~\cite{vatex}} \\
\hline
CLIP~\cite{clip}& 39.7 &72.3& 82.2& 2.0 &52.7 &88.8 &94.9 &1.0\\
SUPPORT~\cite{support} &   44.9& 82.1 &89.7 &\textbf{1.0}&58.4 &84.4 &91.0 &\textbf{1.0} \\
CLIP4Clip~\cite{clip4clip}&  55.9 &89.2& 95.0 &\textbf{1.0}&73.2 &97.1 &99.1 &\textbf{1.0}\\
\rowcolor{Gray}
\textbf{Ours}& \textbf{63.5} & \textbf{91.7} & \textbf{96.5} & \textbf{1.0}  &  \textbf{77.0}  & \textbf{98.0}   & \textbf{99.4}  & \textbf{1.0}\\
\rowcolor{Gray}
\textbf{Ours}$_{\text{+ViT-B/16}}$& 65.7 & 92.6 & 96.7 & 1.0  & 80.1 & 98.5  & 99.5  &  1.0\\
\bottomrule
\multicolumn{9}{c}{Retrieval performance on LSMDC~\cite{lsmdc}} \\
\hline
CE~\cite{ce}&  11.2 &26.9 &34.8& 25.3&- &- &-& - \\
MMT~\cite{mmt}&  12.9 &29.9 &40.1& 19.3&- &- &-& - \\
CLIP~\cite{clip}&   11.3& 22.7& 29.2& 46.5 &- &- &-& - \\
MDMMT~\cite{mmt}&   18.8& 38.5& 47.9&12.3& -& -& -& - \\
CLIP4Clip~\cite{clip4clip}&  22.6& 41.0 &49.1&11.0& -&-&-&- \\
\rowcolor{Gray}
\textbf{Ours}&  \textbf{24.9 }  & \textbf{45.7} & \textbf{55.3}  & \textbf{7.0} & \textbf{24.9}  &\textbf{44.1 } & \textbf{53.8} & \textbf{9.0}  \\
\rowcolor{Gray}
\textbf{Ours}$_{\text{+ViT-B/16}}$&   26.5  &47.6 & 56.8  & 7.0 & 27.0 &45.7 & 55.4 & 8.0  \\
\bottomrule 
\multicolumn{9}{c}{Retrieval performance on ActivityNet~\cite{anet}} \\
\hline
CE~\cite{ce}&   17.7& 46.6& -&6.0&-&-&-&-\\
MMT~\cite{mmt}&   28.9& 61.1 &-&4.0& -& -& -& - \\
SUPPORT\cite{support} &28.7& 60.8& -&\textbf{2.0}&-& -& -& - \\
CLIP4Clip~\cite{clip4clip}&  41.4& 73.7 &85.3& \textbf{2.0}&-&-&-& \\
\rowcolor{Gray}
\textbf{Ours}&  \textbf{44.2} &\textbf{74.5}& \textbf{86.1}& \textbf{2.0} & \textbf{42.2} & \textbf{74.0} &\textbf{86.2} & \textbf{2.0}\\
\rowcolor{Gray}
\textbf{Ours}$_{\text{+ViT-B/16}}$&  46.2  & 77.3 & 88.2 & 2.0 & 45.7  &76.5 & 87.8 & 2.0   \\
\bottomrule
\multicolumn{9}{c}{Retrieval performance on DiDeMo~\cite{didemo}} \\
\hline
CE~\cite{ce}&   15.6& 40.9 &-&8.2 &27.2 &51.7 &62.6& 5.0 \\
ClipBERT~\cite{clipbert}&   21.1& 47.3 &61.1 &6.3& -& -& -& - \\
FROZEN~\cite{frozen} &31.0 &59.8 &72.4& 3.0&-&-&-&-\\
CLIP4Clip~\cite{clip4clip}&  41.4 & 68.2& 79.1 & \textbf{2.0} & 42.8  & 69.8  & 79.0& \textbf{2.0} \\
\rowcolor{Gray}
\textbf{Ours}&  \textbf{47.9}  & \textbf{73.8} & \textbf{82.7} & \textbf{2.0} & \textbf{45.4}  & \textbf{72.6} & \textbf{82.1}  & \textbf{2.0}   \\
\rowcolor{Gray}
\textbf{Ours}$_{\text{+ViT-B/16}}$&   49.0  & 76.5 & 84.5 & 2.0 & 49.9  &75.4 & 83.3 & 2.0   \\
\bottomrule
\end{tabular}
\label{tab:sota_sum}
\vspace{-4mm}
\end{table}
\noindent{disentangled representation method, which are collaboratively implemented by a Weighted Token-wise Interaction (WTI) to solve sequential matching problem from macro-view and a Channel DeCorrelation Regularization (CDCR) that reduces feature redundancy from a micro-view.
We demonstrate the effectiveness of the proposed approach for modeling better sequential and hierarchical clues on six datasets.
We wish our cross-modality interaction will inspire more theoretical researches towards more powerful interaction design.}

\noindent\textbf{Appendix}

\begin{figure}[H]
\centering
\includegraphics[width=0.99\linewidth]{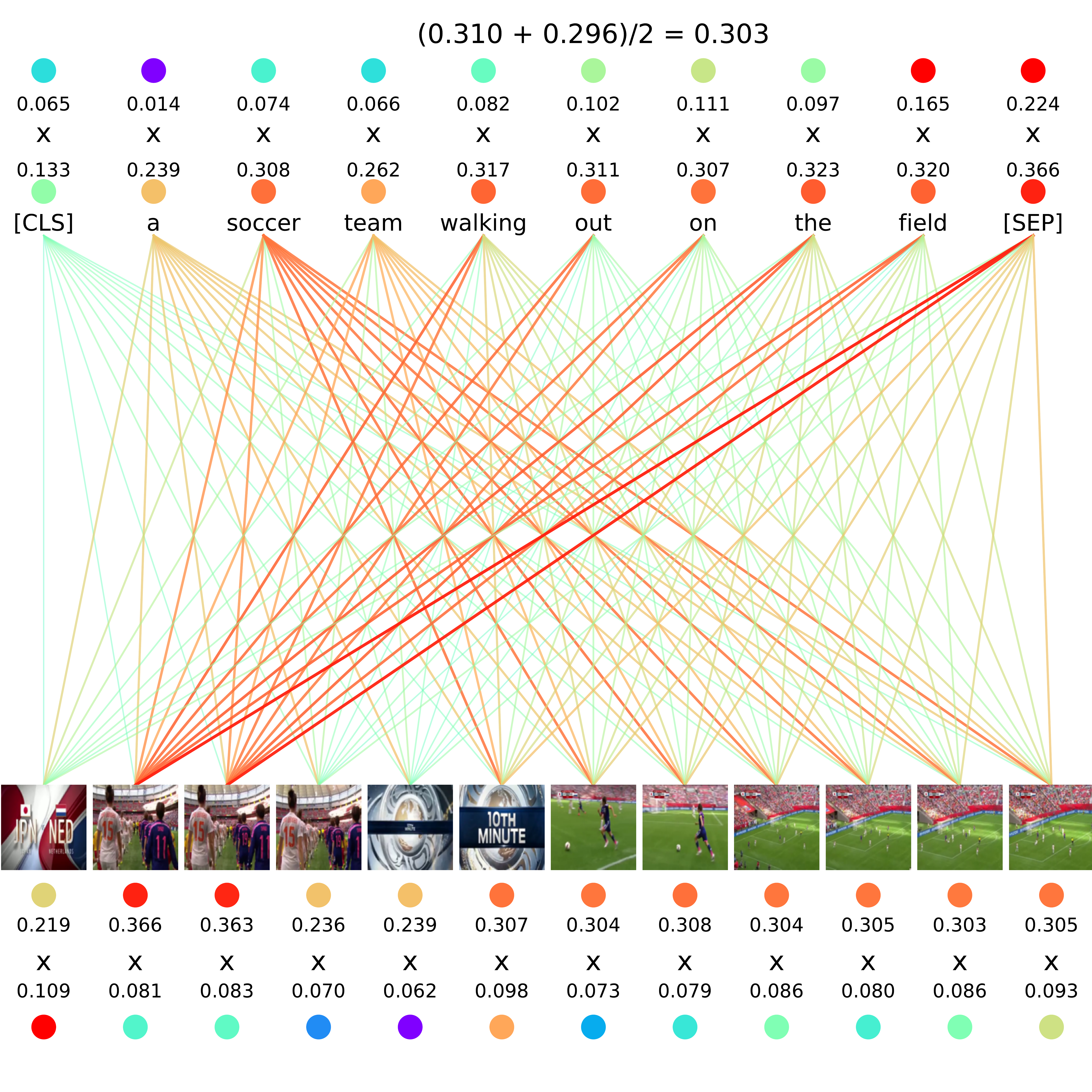}
\caption{Visualization of the internal operating mechanism of WTI on MSR-VTT~\cite{msrvtt}.}
\label{fig:vis}
\end{figure}

This document brings additional details of Weighted Token-wise Interaction. 
Additional qualitative and quantitative results are also given for completeness.
Figure~\ref{fig:vis} shows the internal operating mechanism of WTI, and also powerfully explains the performance improvement of our algorithm.

\begin{figure}[t]
\centering
\includegraphics[width=0.94\linewidth]{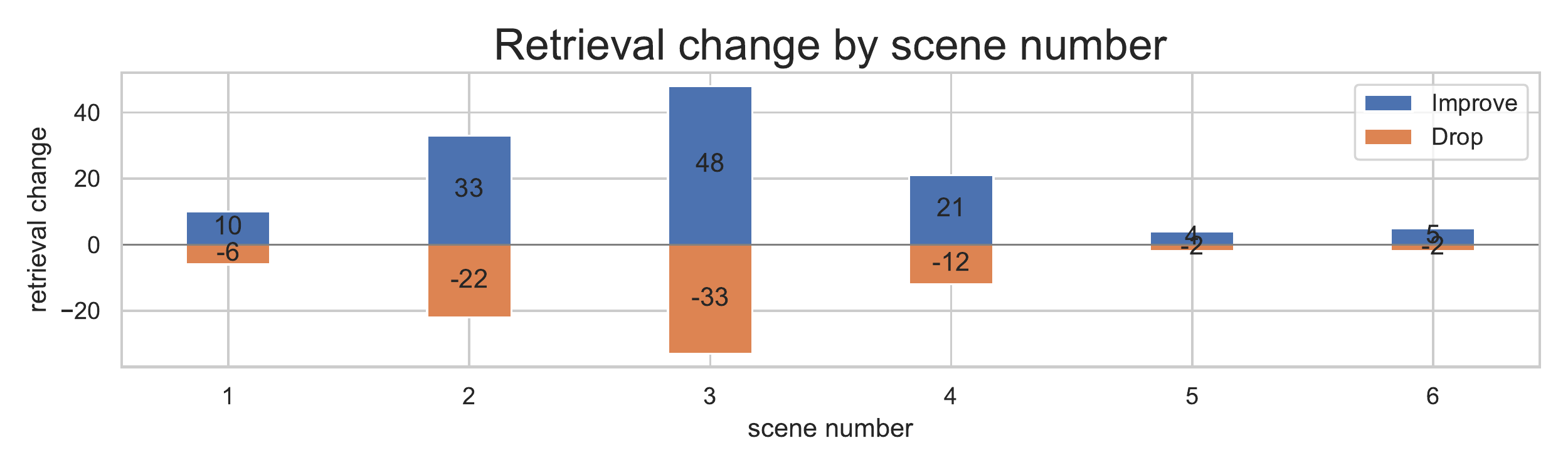}
\vspace{-6mm}
\caption{The improvement with different scenes number on the MSR-VTT 1K val~\cite{msrvtt}.}
\label{fig:scene}
\end{figure}
\begin{figure}[t]
\centering
\includegraphics[width=0.94\linewidth]{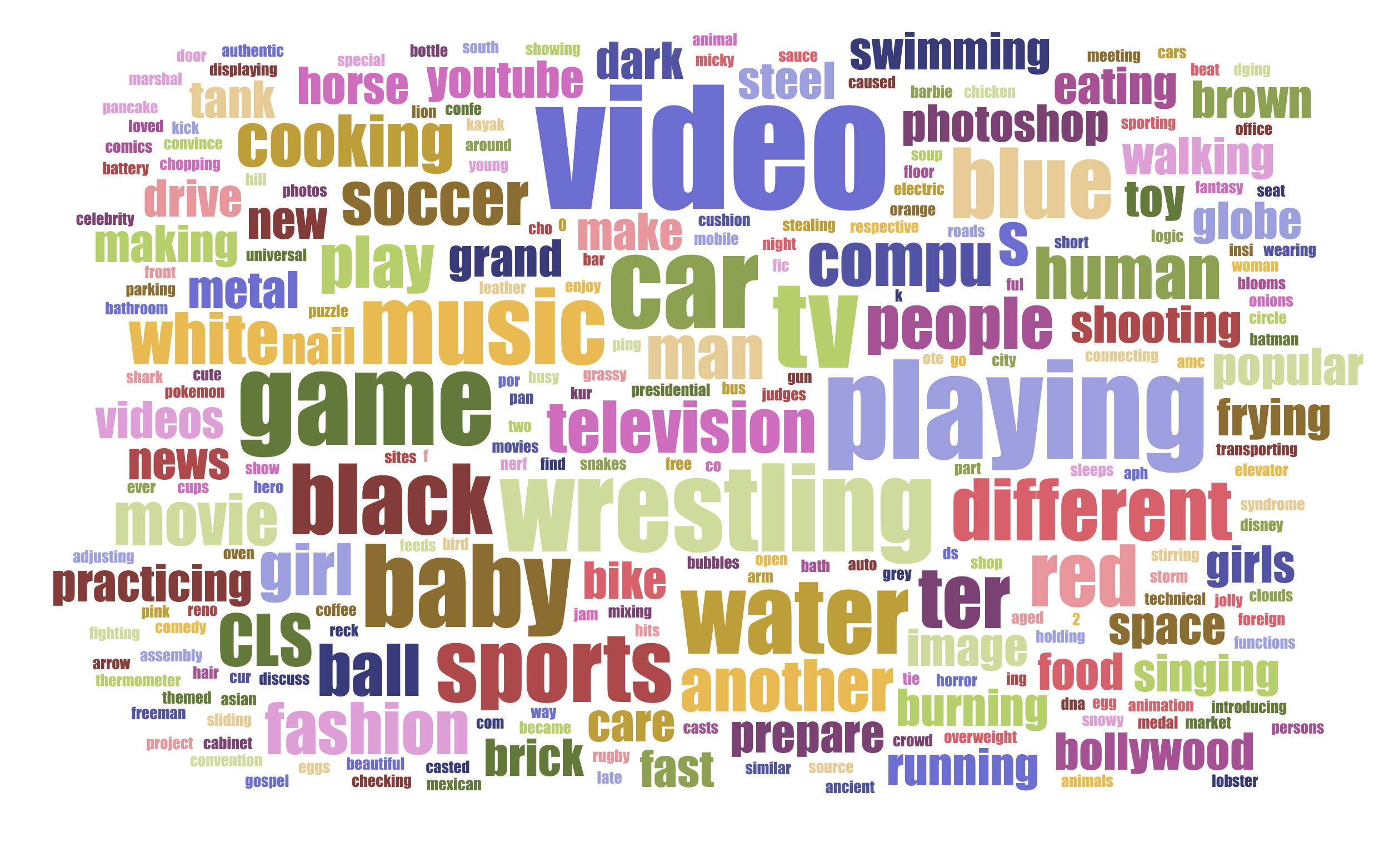}
\vspace{-6mm}
\caption{The word cloud of top-1 weighted ordinary tokens on the MSR-VTT 1K val~\cite{msrvtt}.}
\label{fig:word}
\end{figure}

\section{Discussion}
\textbf{How Weighted Token-wise Interaction (WTI) improve performance?} 
In this section, we discuss the reasons for considering \textbf{token-wise} interaction and the \textbf{adaptive weight} module. 

First, the text and video input are naturally sequential, and keeping the sequential structure can reduce the information loss and preserve more fine-grained cues.
Using the sequential representation for interaction is essentially an ensemble learning method.
The \textbf{Max} operation in our token-wise interaction can select matching video clips, as shown in Fig.~\ref{fig:vis}, which can help us resolve \textbf{local context assumptions} in the main paper.
Otherwise, if using average pooling ~\cite{clip4clip}, the overall similarity will be pulled down by irrelevant segments.
Quantitatively, we analyze the relationship between recall improvement and the number of scenes in a video (Fig.~\ref{fig:scene}), and find that our algorithm achieves most improvements at 3 scenes.
Therefore, token-wise interaction is intuitively a sensible practice. 

Second, learning of adaptive weights is necessary, especially when CLIP~\cite{clip} is used as a pre-trained network.
CLIP performs large-scale training (on 256 GPUs for 2 weeks) with large-scale dataset (400 millions), and we cannot completely change the distribution of position embedding and features by a small-scale fine-tuning dataset.
The default global token, \texttt{[SEP]}, shows a dominant role, achieving top1 important token for $98.7\%$ videos and an average weight of 0.2838, far exceeding the average weight of other tokens (0.0674) on MSR-VTT 1K validation set.
Simply averaging all sequential features is unfair to \texttt{[SEP]}, resulting in a suboptimal solution.
On the other hand, the importance of words in the text are also quite different, and our experimental analysis shows that nouns/verbs account for the main weight, as shown in Fig.~\ref{fig:word}.

Therefore, a good practice is to consider the token-wise interaction and adaptive weight module and devising a solution to effectively fuse these two factors (as WTI).

\section{More Implementation Details}
\label{sec:details}
In this section, we present more implementation details that were omitted in the main paper for brevity.

\subsection{Pseudocode with Mask Inputs}
Please note we omit the \textbf{mask} inputs at the main paper for brevity in Section 3.2. 
Here we present complete pseudo-codes for our Weighted Token-wise Interaction (WTI) and Channel Decorrelation Regularization (CDCR) in Algorithm~\ref{alg:pytorchmask}-~\ref{alg:cdcr2}.
We believe the pseudocode would aid an independent researcher to better replicate the proposed interaction. 

\begin{algorithm}[tb]
\caption{PyTorch-style pseudocode for Weighted Token-wise Interaction.}
\label{alg:pytorchmask}
   
    \definecolor{codeblue}{rgb}{0.25,0.5,0.5}
    \definecolor{deepred}{rgb}{0.95,0,0}
    \lstset{
      basicstyle=\fontsize{7.2pt}{7.2pt}\ttfamily\bfseries,
      commentstyle=\fontsize{7.2pt}{7.2pt}\color{codeblue},
      keywordstyle=\fontsize{7.2pt}{7.2pt},
      emph={mask_t, mask_v,v2t_max_idx,t2v_max_idx},          %
      emphstyle=\color{red},    %
    }
\begin{lstlisting}[language=python]
# t: text input                 v: video input
# mask_t: text mask             mask_v: video mask
# f_t: text encoder network     f_v: video encoder network
# f_tw: text weight network     f_vw: video weight network
# B: batch size                 D: dimensionality of the embeddings
# N_t: text token size          N_v: frame token size

def weighted_token_wise_interaction(t, mask_t, v, mask_v):    
    # compute embeddings
    e_t = f_t(t) # BxN_txD
    e_v = f_v(v) # BxN_vxD
    
    # generate fusion weights
    text_weight = f_tw(e_t).squeeze()  # BxN_t
    # fill masked text with -inf
    text_weight.masked_fill_(1 - mask_t, float("-inf"))  # BxN_v
    text_weight = torch.softmax(text_weight, dim=-1) # BxN_t
    
    video_weight = f_tv(e_v).squeeze()  # BxN_v
    # fill masked video with -inf
    vision_weight.masked_fill_(1 - mask_v, float("-inf"))  # BxN_v
    video_weight = torch.softmax(vision_weight, dim=-1)    # BxN_v
    
    # normalize representation
    e_t = e_t / e_t.norm(dim=-1, keepdim=True)  # BxN_txD
    e_v = e_v / e_v.norm(dim=-1, keepdim=True)  # BxN_vxD
    
    # token interaction
    logits = torch.einsum("atc,bvc->abtv", [e_t, e_v])  # BxBxN_txN_v

    # mask for logits
    logits = torch.einsum('abtv,at->abtv', [logits, mask_t])
    logits = torch.einsum('abtv,bv->abtv', [logits, mask_v])    
    
    t2v_logits, t2v_max_idx = logits.max(dim=-1)  # BxBxN_txN_v -> BxBxN_t
    t2v_logits = torch.einsum("abt,at->ab", [t2v_logits, text_weight])  # BxBxN_t -> BxB
    v2t_logits, v2t_max_idx = logits.max(dim=-2)  # BxBxN_txN_v -> BxBxN_v
    v2t_logits = torch.einsum("abv,bv->ab", [v2t_logits, video_weight])  # BxBxN_v-> BxB
    
    retrieval_logits = (t2v_logits + v2t_logits) / 2.0  # BxB
    
    return retrieval_logits
\end{lstlisting}
\end{algorithm}

\begin{algorithm}[tb]
\caption{PyTorch-style pseudocode for Channel DeCorrelation Regularization with single-vector representation.}
\label{alg:cdcr}
   
    \definecolor{codeblue}{rgb}{0.25,0.5,0.5}
    \definecolor{deepred}{rgb}{0.95,0,0}
    \lstset{
      basicstyle=\fontsize{7.2pt}{7.2pt}\ttfamily\bfseries,
      commentstyle=\fontsize{7.2pt}{7.2pt}\color{codeblue},
      keywordstyle=\fontsize{7.2pt}{7.2pt},
      emph={mask_t, mask_v,__init__},          %
      emphstyle=\color{red},    %
    }
\begin{lstlisting}[language=python]
# e_t: text feature              e_v: video feature
# B: batch size                 D: dimensionality of the embeddings
# alpha: the magnitude of the redundancy term

def channel_decorrelation_regularization_single(e_t, e_v):    
    # batch norm
    t_norm = (e_t - e_t.mean(0)) / e_t.std(0)  # NxD
    v_norm = (e_v - e_v.mean(0)) / e_v.std(0)  # NxD
    
    B, D = t_norm.shape
    cov = torch.einsum('ac,ad->cd', t_norm, v_norm) / B  # DxD
    # loss
    on_diag = torch.diagonal(cov).add_(-1).pow_(2).sum()
    off_diag = cov.flatten()[1:].view(D - 1, D + 1)[:, :-1].pow_(2).sum()
    cdcr = on_diag + off_diag * alpha
    return cdcr
\end{lstlisting}
\end{algorithm}

\begin{algorithm}[tb]
\caption{PyTorch-style pseudocode for Channel DeCorrelation Regularization with sequential representation.}
\label{alg:cdcr2}
   
    \definecolor{codeblue}{rgb}{0.25,0.5,0.5}
    \definecolor{deepred}{rgb}{0.95,0,0}
    \lstset{
      basicstyle=\fontsize{7.2pt}{7.2pt}\ttfamily\bfseries,
      commentstyle=\fontsize{7.2pt}{7.2pt}\color{codeblue},
      keywordstyle=\fontsize{7.2pt}{7.2pt},
      emph={mask_t, mask_v,v2t_max_idx,t2v_max_idx},          %
      emphstyle=\color{red},    %
    }
\begin{lstlisting}[language=python]
# e_t: text feature            e_v: video feature
# mask_t: text mask             mask_v: video mask
# B: batch size                 D: dimensionality of the embeddings
# N_t: text token size          N_v: frame token size

def channel_decorrelation_regularization_sequential(e_t, mask_t, e_v, mask_v): 
    # selecet max indexs for each text-video pair
    i4t = t2v_max_idx[torch.arange(B),torch.arange(B)]  # BxBxN_t -> BxN_t 
    i4v = v2t_max_idx[torch.arange(B), torch.arange(B)]  # BxBxN_v -> BxN_v
    
    e_4_t = e_v[torch.arange(B).repeat_interleave(N_t), i4t.flatten()]  # (BxN_t)xD
    e_4_v = e_t[torch.arange(B).repeat_interleave(N_v), i4v.flatten()]  # (BxN_v)xD
    e_t = e_t.reshape(-1,D)  # (BxN_t)xD
    e_v = e_v.reshape(-1,D)  # (BxN_v)xD
    mask_t = mask_t.flatten().type(torch.bool)
    mask_v = mask_v.flatten().type(torch.bool)
    
    e_t = e_t[mask_t]
    e_v = e_v[mask_v]
    e_4_t = e_4_t[mask_t]
    e_4_v = e_4_v[mask_v]
    
    # cov for t2v
    t_norm = (e_t - e_t.mean(0)) / e_t.std(0)  # XxD
    v_norm = (e_4_t - e_4_t.mean(0)) / e_4_t.std(0)  # XxD
    X = t_norm.shape
    cov1 = torch.einsum('ac,ad->cd', t_norm, v_norm) / B  # DxD
    
    # cov for v2t
    v_norm = (e_t - e_t.mean(0)) / e_t.std(0)  # XxD
    t_norm = (e_4_v - e_4_v.mean(0)) / e_4_v.std(0)  # XxD
    X= t_norm.shape[0]
    cov2 = torch.einsum('ac,ad->cd', t_norm, v_norm) / B  # DxD
    
    cov = (cov1+cov2)/2
    
    # loss
    on_diag = torch.diagonal(cov).add_(-1).pow_(2).sum()
    off_diag = cov.flatten()[1:].view(D - 1, D + 1)[:, :-1].pow_(2).sum()
    cdcr = on_diag + off_diag * alpha
    return cdcr
\end{lstlisting}
\end{algorithm}

\subsection{Network Architecture for Weight Branch}

Figure~\ref{fig:vis} shows details of the network architectures for learning weights. 
We use the single-modal input for each branch to avoid coupling between modalities.

\begin{figure}[h]
\centering
\centering
\includegraphics[width=0.9\linewidth]{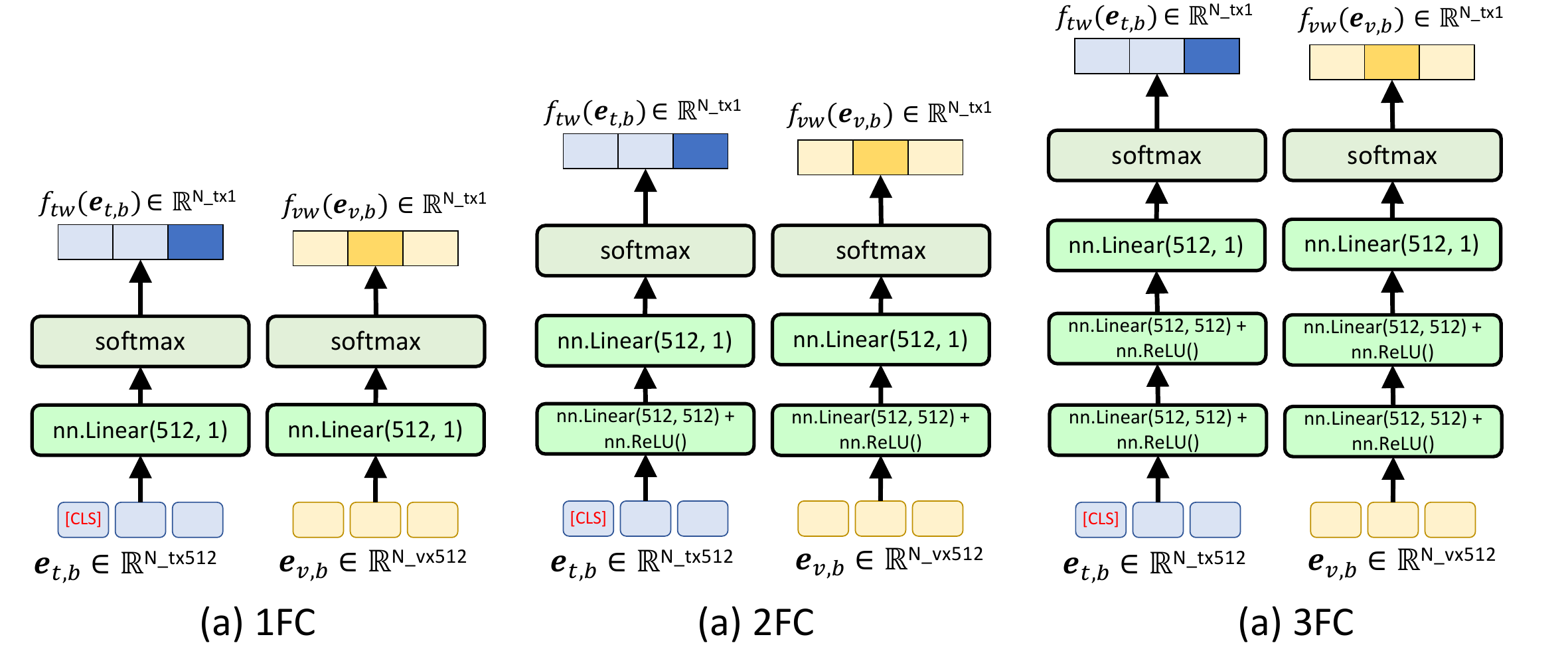}
\caption{Network architectures for weight branch.}
\label{fig:weight}
\end{figure}

\subsection{Complexity Analysis}

\begin{table}[t]
\centering
\caption{Comparisons of different interaction mechanisms. $N$ denotes number of video documents ($N$ is large and depends on applications); $D$ denotes representation dimension ($D=512$, by default); $N_t$ and $N_v$ denote length of text token and frame token, $N_{v+t}$ is the sum of $N_t$ and $N_v$ ($N_t = 32, N_v = 12$); $L$ denotes number of network layer and $S$ denotes feature levels ($L=4, S=3$, by default).}
\begin{tabular}{l|c|c| c|c}
\toprule
Interaction & Contents    & Function          & Computational Complexity           & Memory   \\
\hline
DP   & single  & parameter-free       & $\mathcal{O}(ND)$              & $\mathcal{O}(ND)$  \\
HI                   & multi-level    & light-parameter & $\mathcal{O}(NSD)$             & $\mathcal{O}(NSD)$ \\
MLP                   & single    & black-box            & $\mathcal{O}(N(D^2L+D))$        & $\mathcal{O}(ND)$  \\
XTI      & token-wise   & black-box            & $\mathcal{O}(N(D^2 N_{t+v} + N_{t+v}^2D)L)$ & $\mathcal{O}(NN_{v}D)$ \\
TI                    & token-wise   & parameter-free       & $\mathcal{O}(NN_tN_vD)$            & $\mathcal{O}(NN_{v}D)$ \\
WTI                   & token-wise   & light-parameter & $\mathcal{O}(N(N_tN_vD+N_{t+v}))$            & $\mathcal{O}(NN_{v}(D+1))$\\
\bottomrule
\end{tabular}
\label{tab:cmp2}
\end{table}%

\noindent{\textbf{Single Vector Dot-product Interaction:}}
For each video, we only need to store a $D$-dimensional feature, and the the overall feature storage complexity is $\mathcal{O}(ND)$.
During the query process, a similarity calculation consists of $D$ multiplication operations.
The overall computational complexity is $\mathcal{O}(ND)$ .

\noindent{\textbf{Hierarchical Interaction:}}
For the hierarchical interaction, we consider the simplest case, directly using the constant weighted average to fuse the multi-layer features.
Therefore, both computation and storage are increased by a factor of $S$ compared to DP.

\noindent{\textbf{MLP on Global Vector:}}
We adopt the structure of stacking $L\times$ FC+ReLU blocks, where the input channel of the first layer is $2D$, the hidden channels size is $D$, and the final output is a single similarity.
The computational complexity is  $\mathcal{O}(N(2DD+DD(L-2)+D)) = \mathcal{O}(N(LD^2+D))$.
We can observe that with the default configuration ($D=512, L=4$), the computational complexity of MLP is about \textbf{2,049}$\times$ of the DP, which is why neural network-based interactions are difficult to deploy. 
They are usually designed to be used as a fine-grained ranking module.
The storage complexity of MLP is same as DP.

\noindent{\textbf{Cross Transformer Interaction:}}
It is usually assumed that a Multi-Head Attention block (MHA) consists of linear layer and \textbf{QKV} structure. The length of the sequence input is $N_{t+v}$, so the computational complexity of the linear layer is $\mathcal{O}(D^2N_{t+v})$.
An dot-product version of the QKV operation requires $\mathcal{O}(N_{t+v}^2D)$ complexity.
Therefore, the overall computational complexity for $L\times$ MHA is $\mathcal{O}(N(D^2 N_{t+v} + N_{t+v}^2D)L)$.
With the default configuration ($D=512, L=4, N_t=32, N_v=12, N_{t+v}=44$), the computational complexity of XTI  is about \textbf{24,464}$\times$ of the DP.
At the same time, due to the use of sequential representation for video, the storage complexity rises to $\mathcal{O}(NN_{v}D)$.

\noindent{\textbf{Token-wise Interaction:}}
Token-wise interactions need to calculate an $N_t\times N_v$ size of similarity matrix for text and video tokens.
The computational complexity is $\mathcal{O}(N(N_tN_vD))$ and the storage complexity is $\mathcal{O}(NN_{v}D)$.

\noindent{\textbf{Weighted Token-wise Interaction:}}
For weighted token-wise interaction, the complexity for the online search process is a little more complicated.
For the query sentence, we need to calculate an additional MLP to generate the text weights, which brings $\mathcal{O}(D^2LN_s)$.
For video documents, we can \textbf{pre-compute} sequential video features and the corresponding video weights.
The total storage overhead is $\mathcal{O}(N(N_vD+N_{v}))$.
Note that \textbf{during the online process, there is no need to dynamically calculate the weights for each video.}
Compared with TI, only two additional weighting operations are added in the online process. Therefore, the overall computational complexity is $\mathcal{O}(N(N_tN_vD+N_{t+v}))$.
Compared to XTI, our computational complexity is reduced \textbf{63.69} times.

\section{Visualizations}

\begin{figure}[h]
\centering
\includegraphics[width=0.4\linewidth]{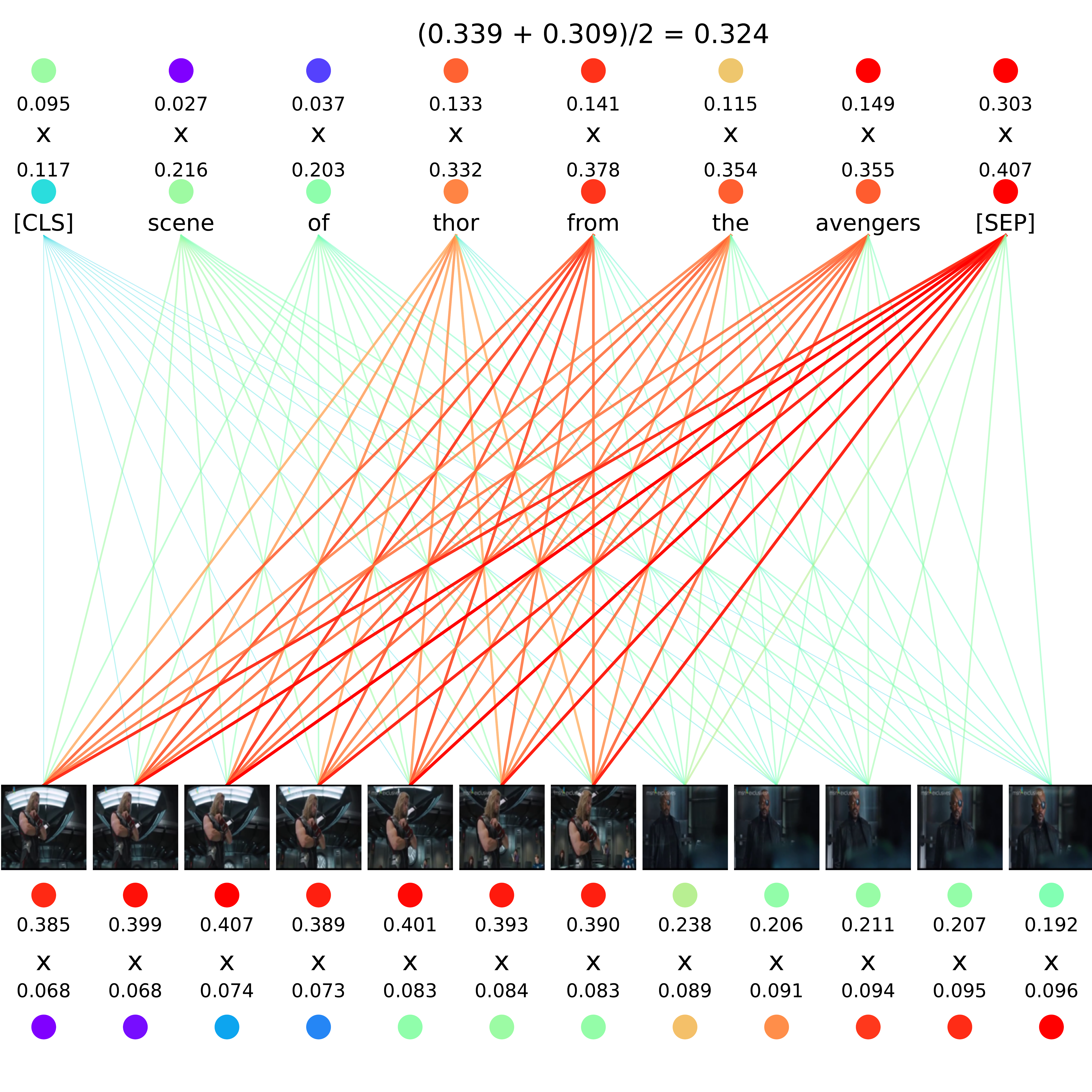}
\includegraphics[width=0.4\linewidth]{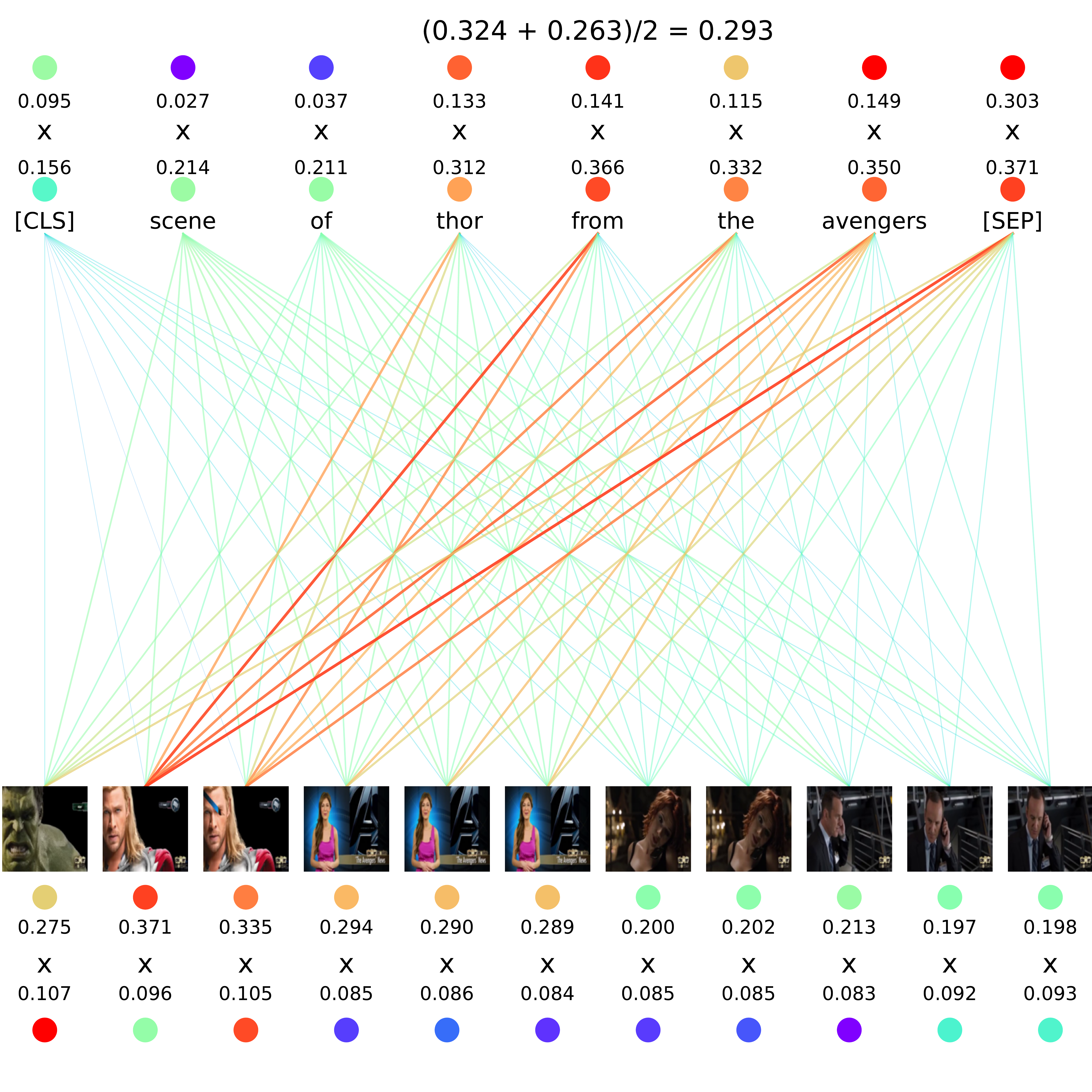}
\includegraphics[width=0.4\linewidth]{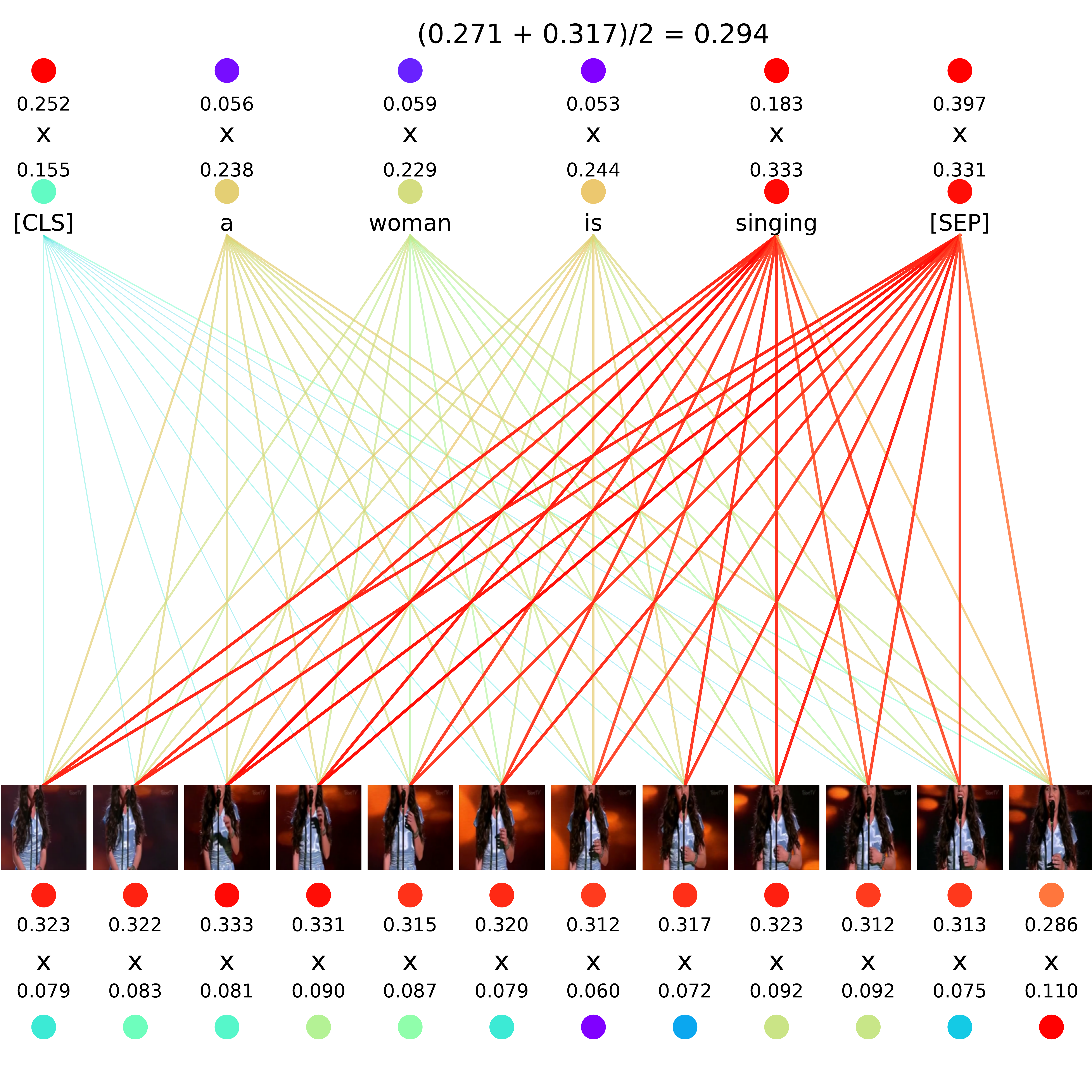}
\includegraphics[width=0.4\linewidth]{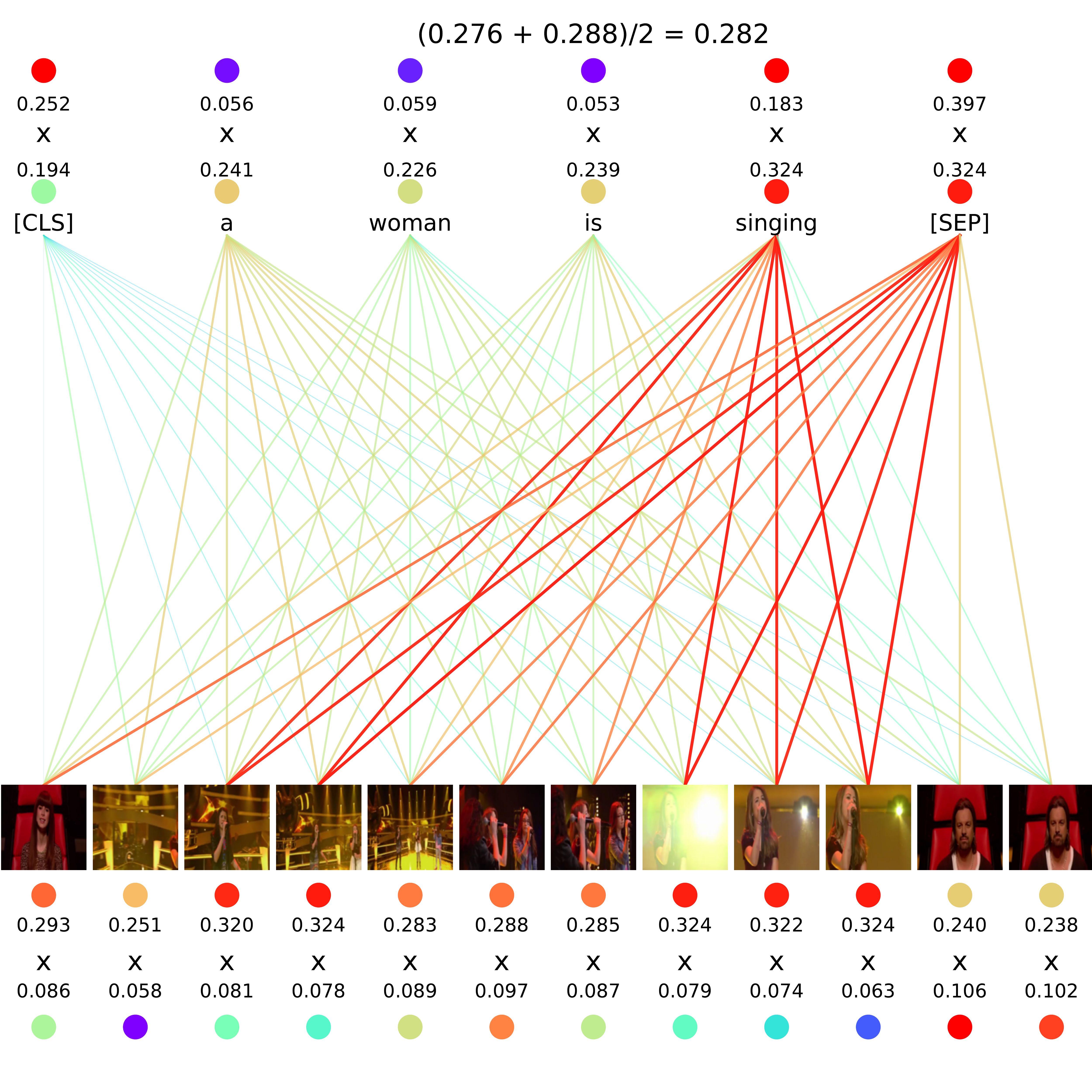}
\includegraphics[width=0.4\linewidth]{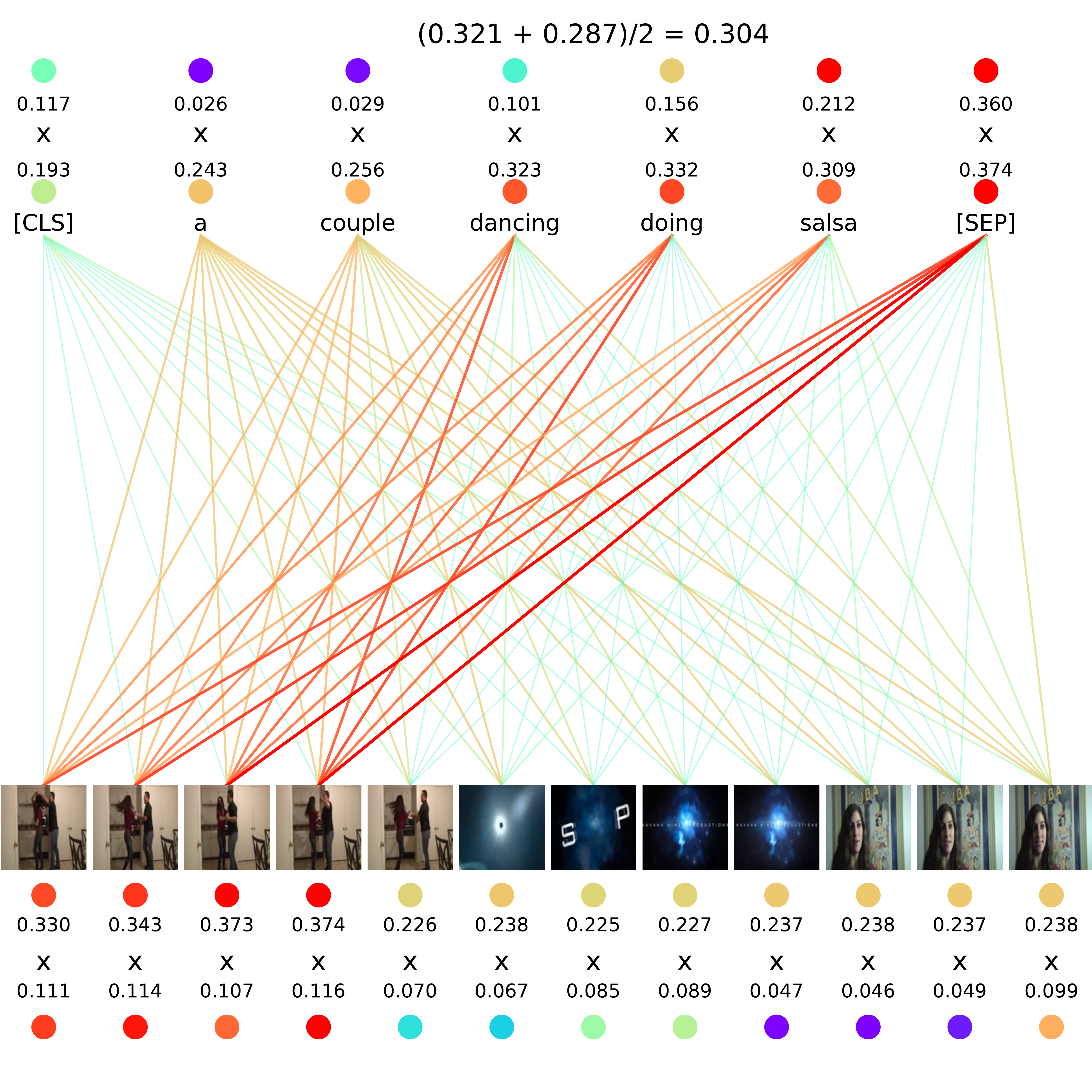}
\includegraphics[width=0.4\linewidth]{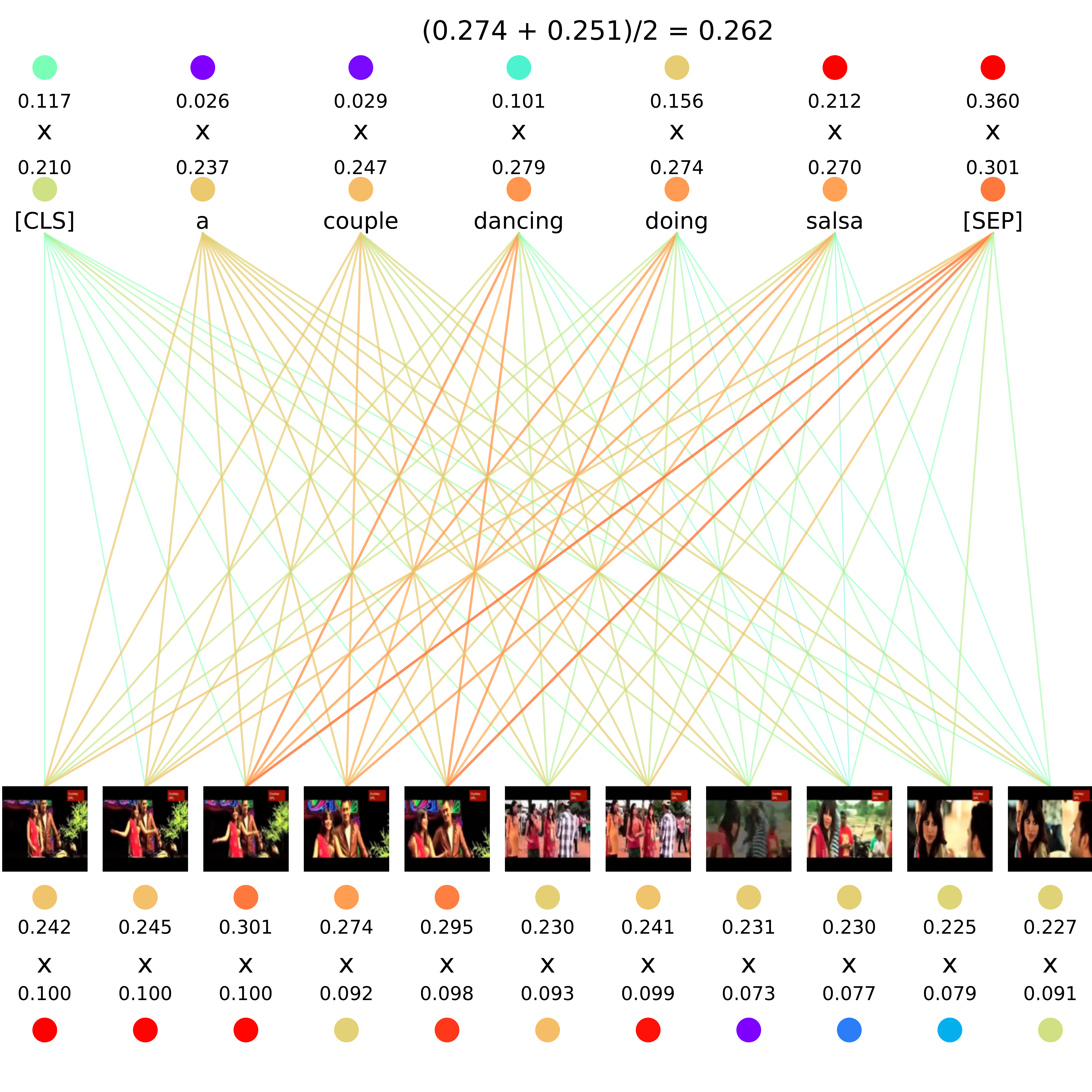}
\includegraphics[width=0.4\linewidth]{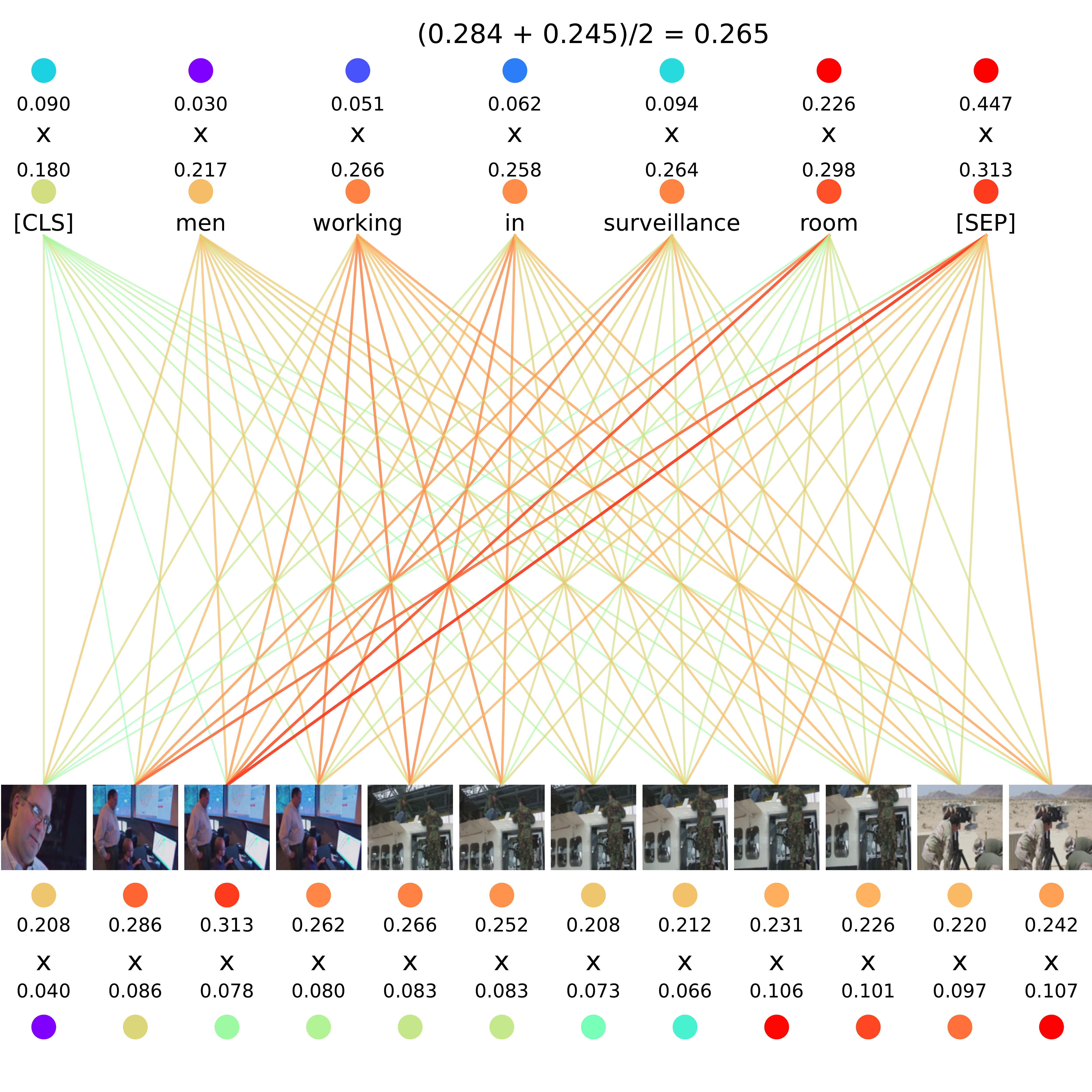}
\includegraphics[width=0.4\linewidth]{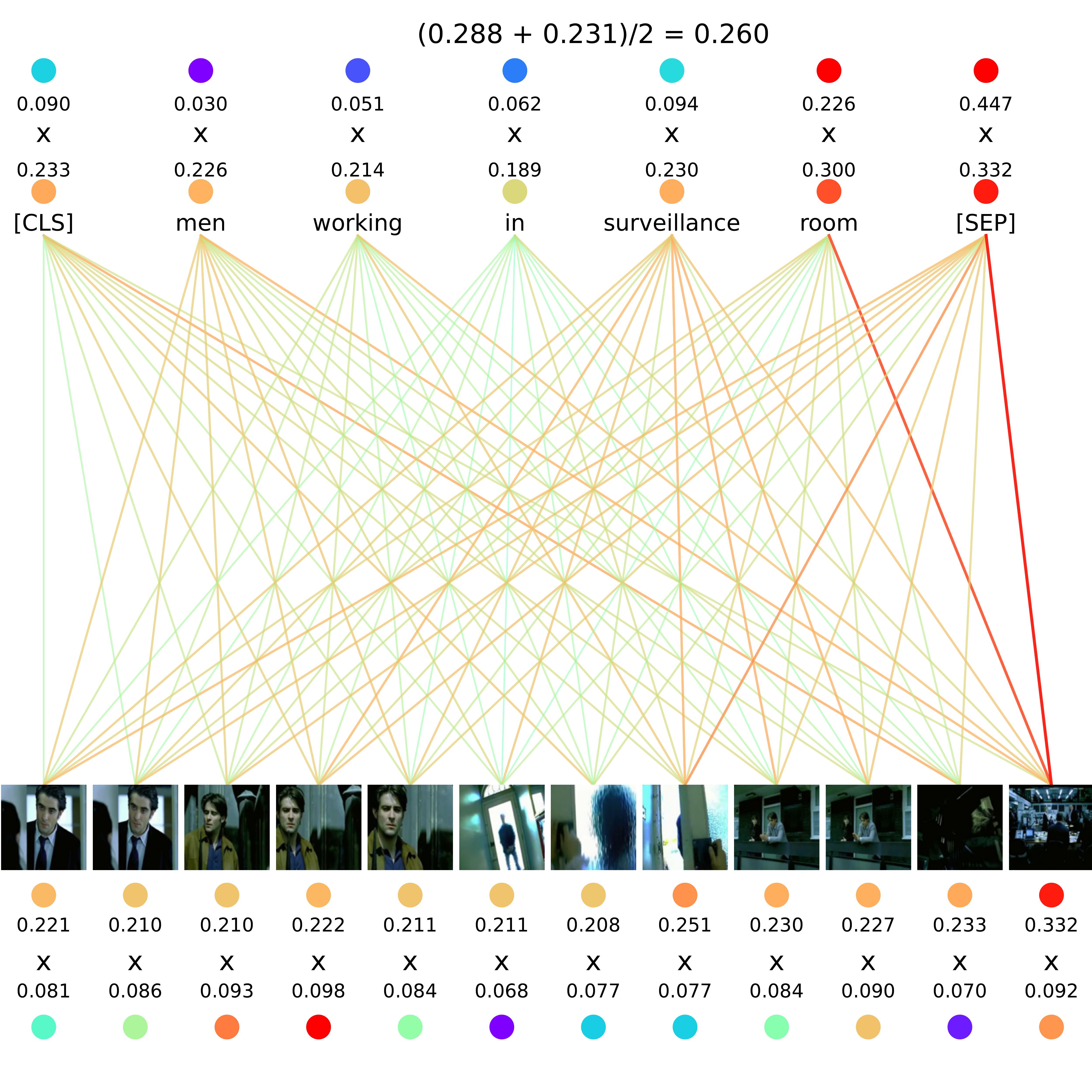}
\caption{Qualitative results on images from MSR-VTT~\cite{msrvtt}.}
\label{fig:msrvtt1}
\end{figure}

\begin{figure}[h]
\centering
\includegraphics[width=0.4\linewidth]{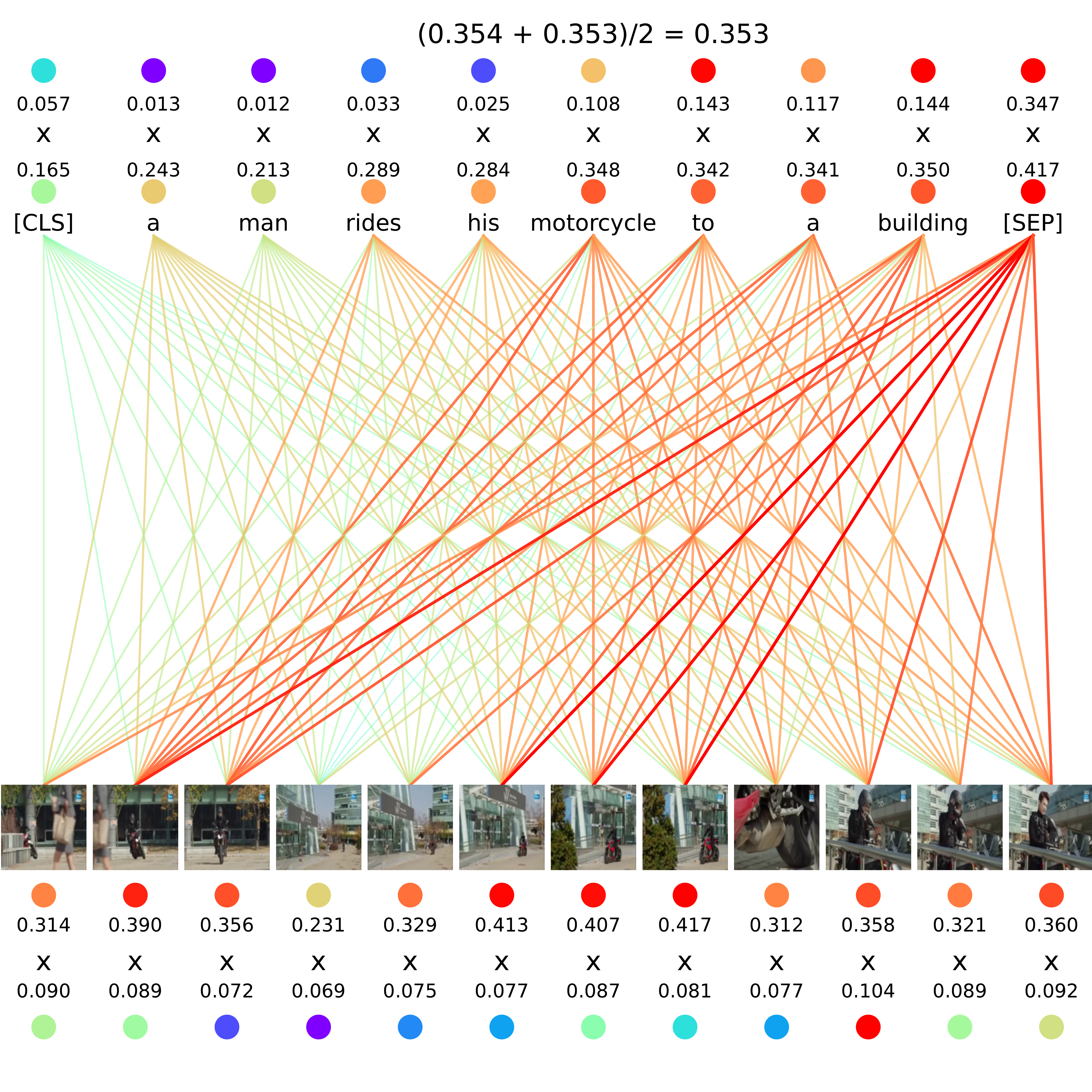}
\includegraphics[width=0.4\linewidth]{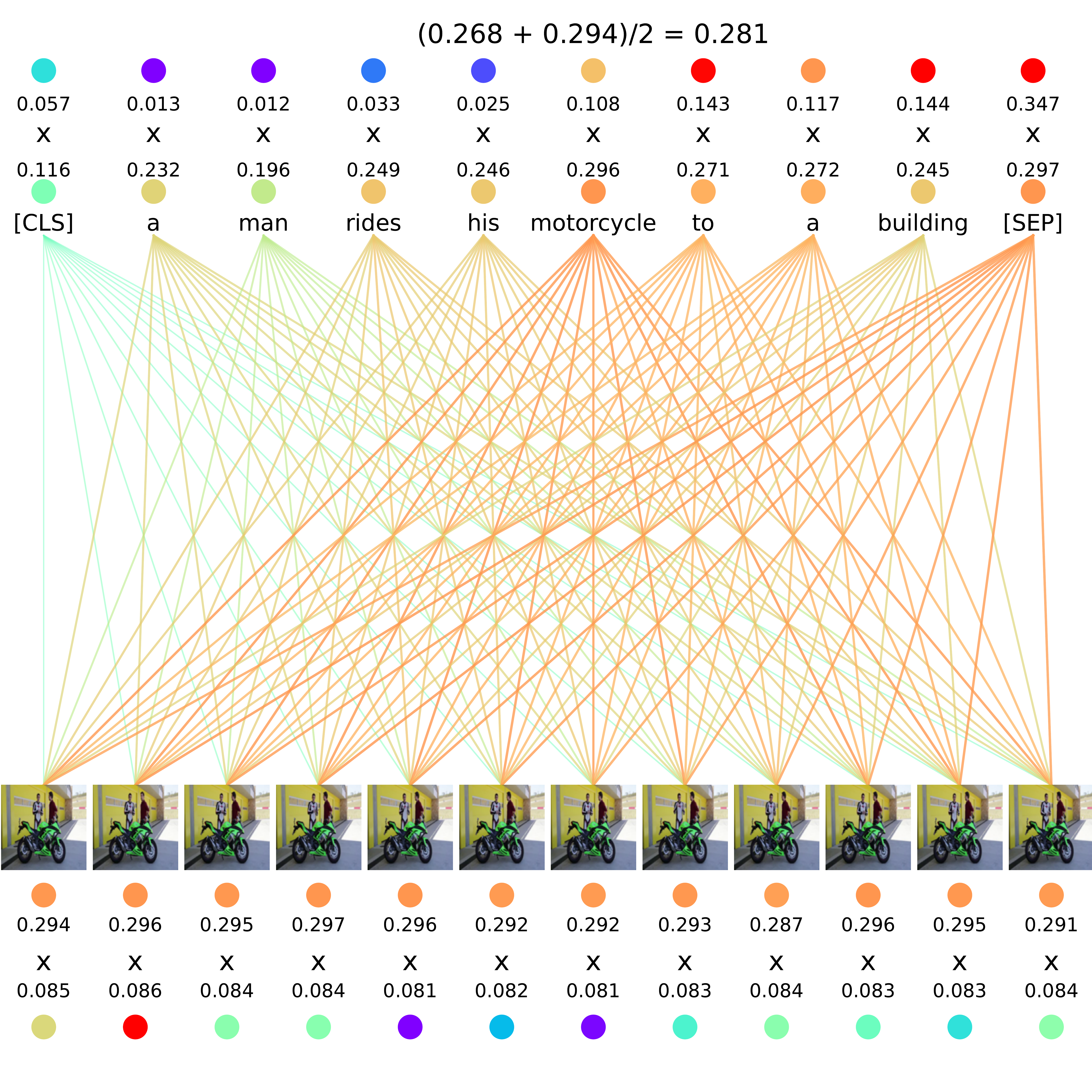}
\includegraphics[width=0.4\linewidth]{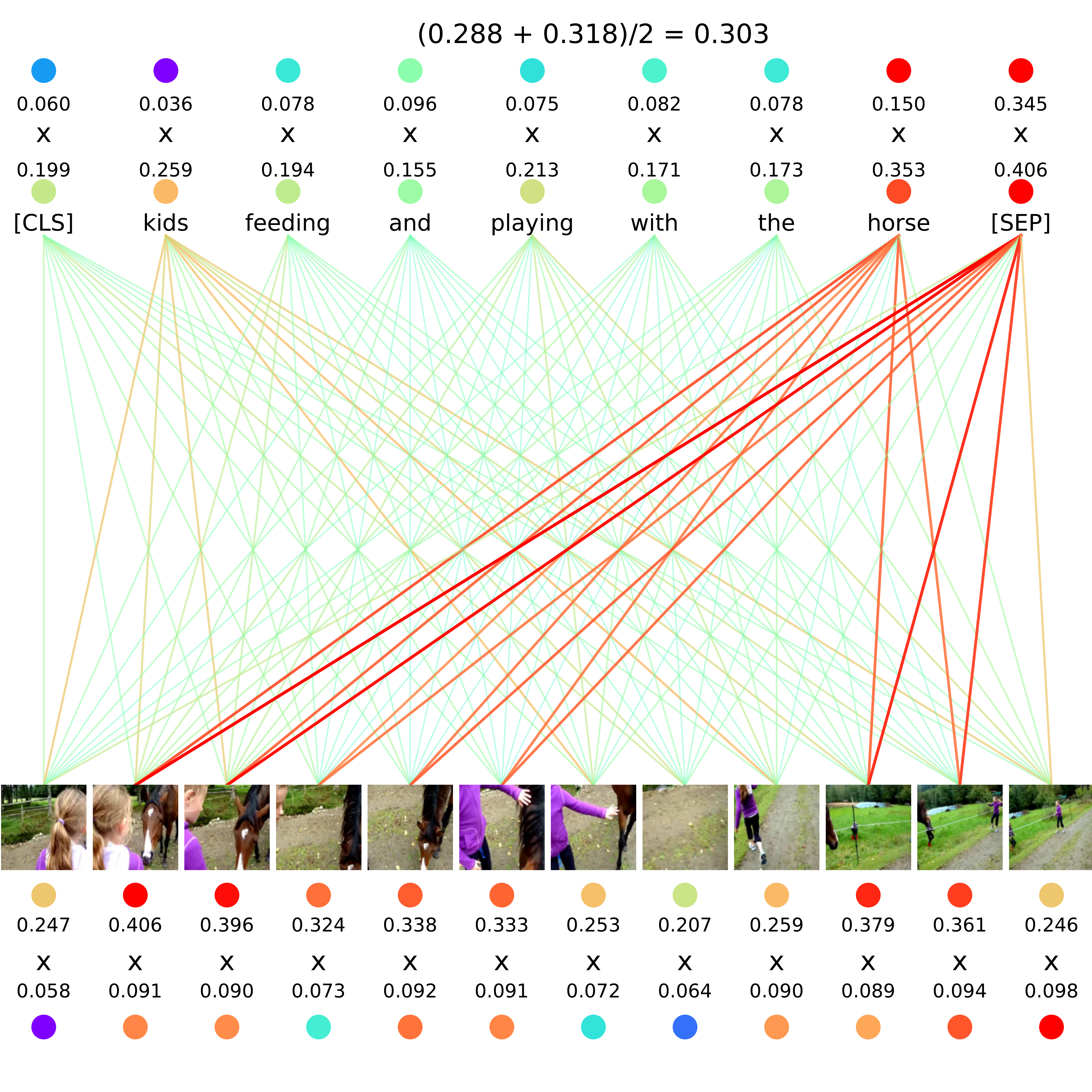}
\includegraphics[width=0.4\linewidth]{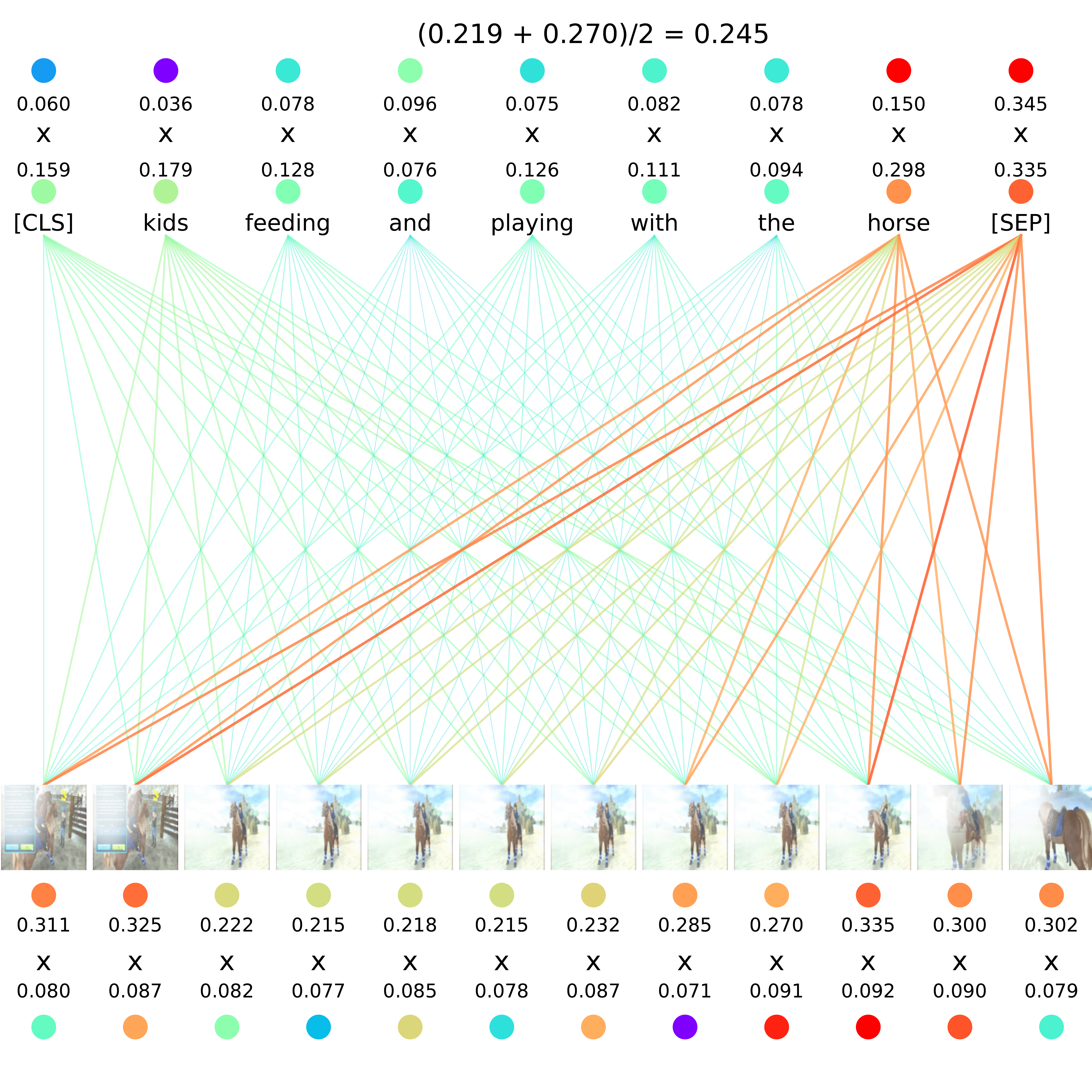}
\includegraphics[width=0.4\linewidth]{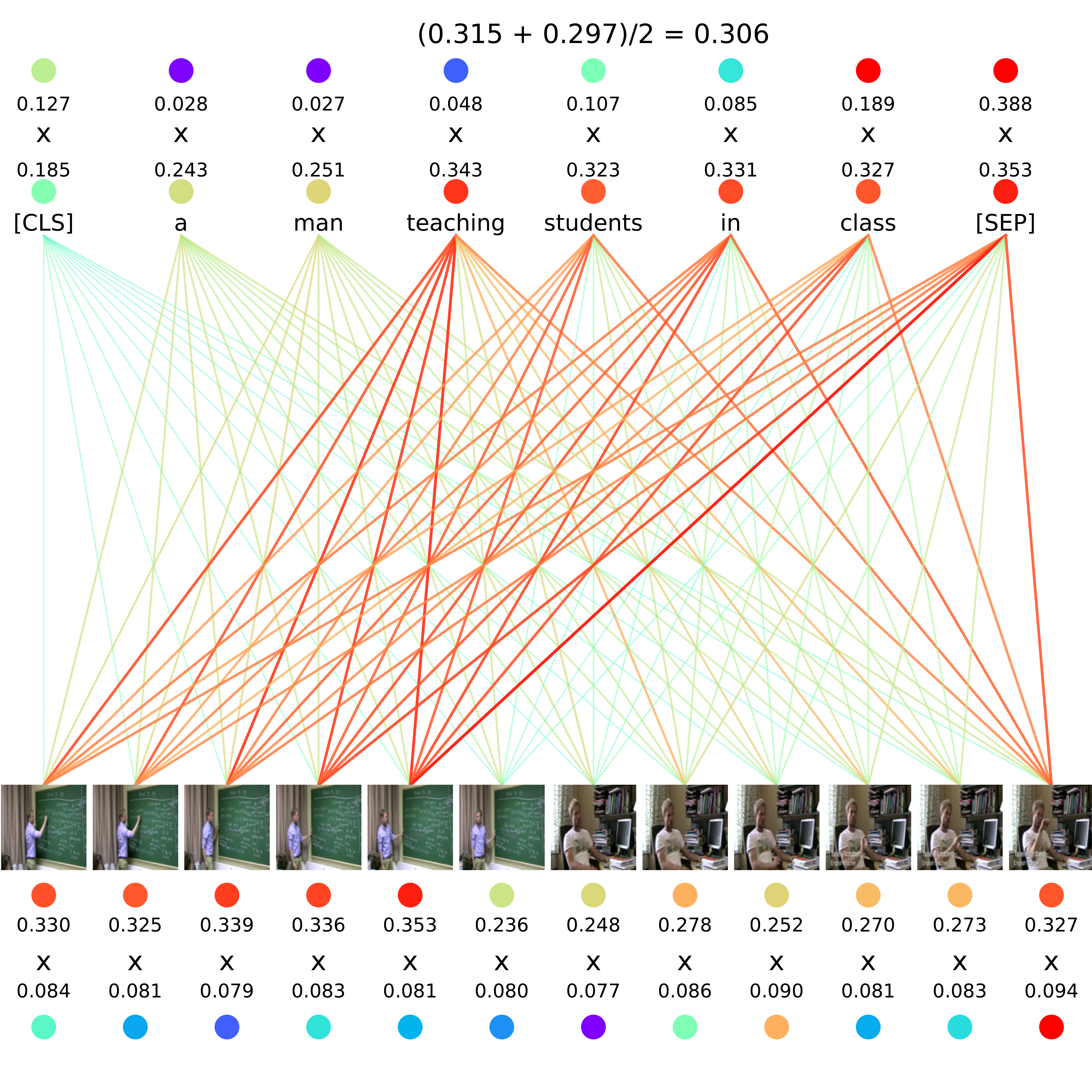}
\includegraphics[width=0.4\linewidth]{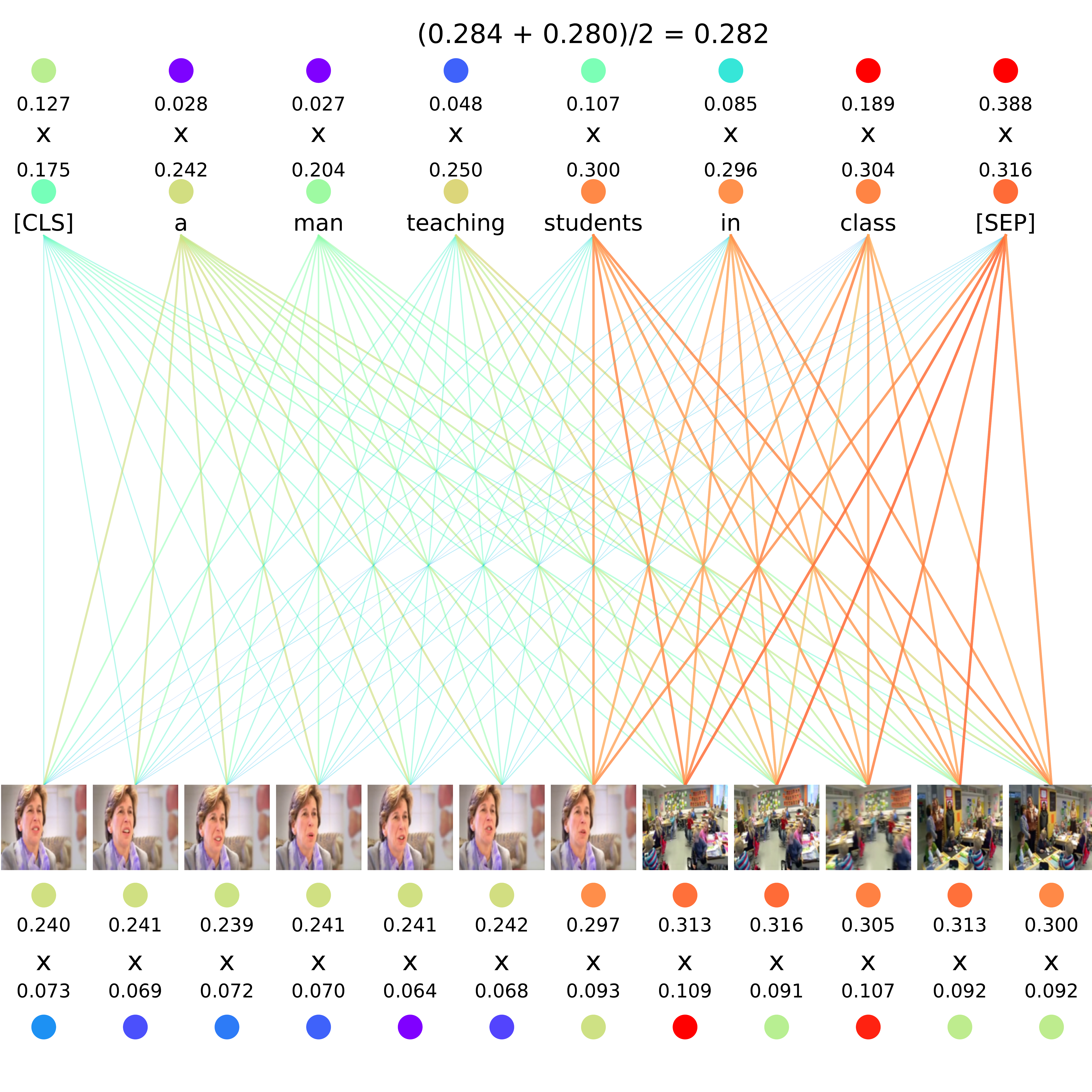}
\includegraphics[width=0.4\linewidth]{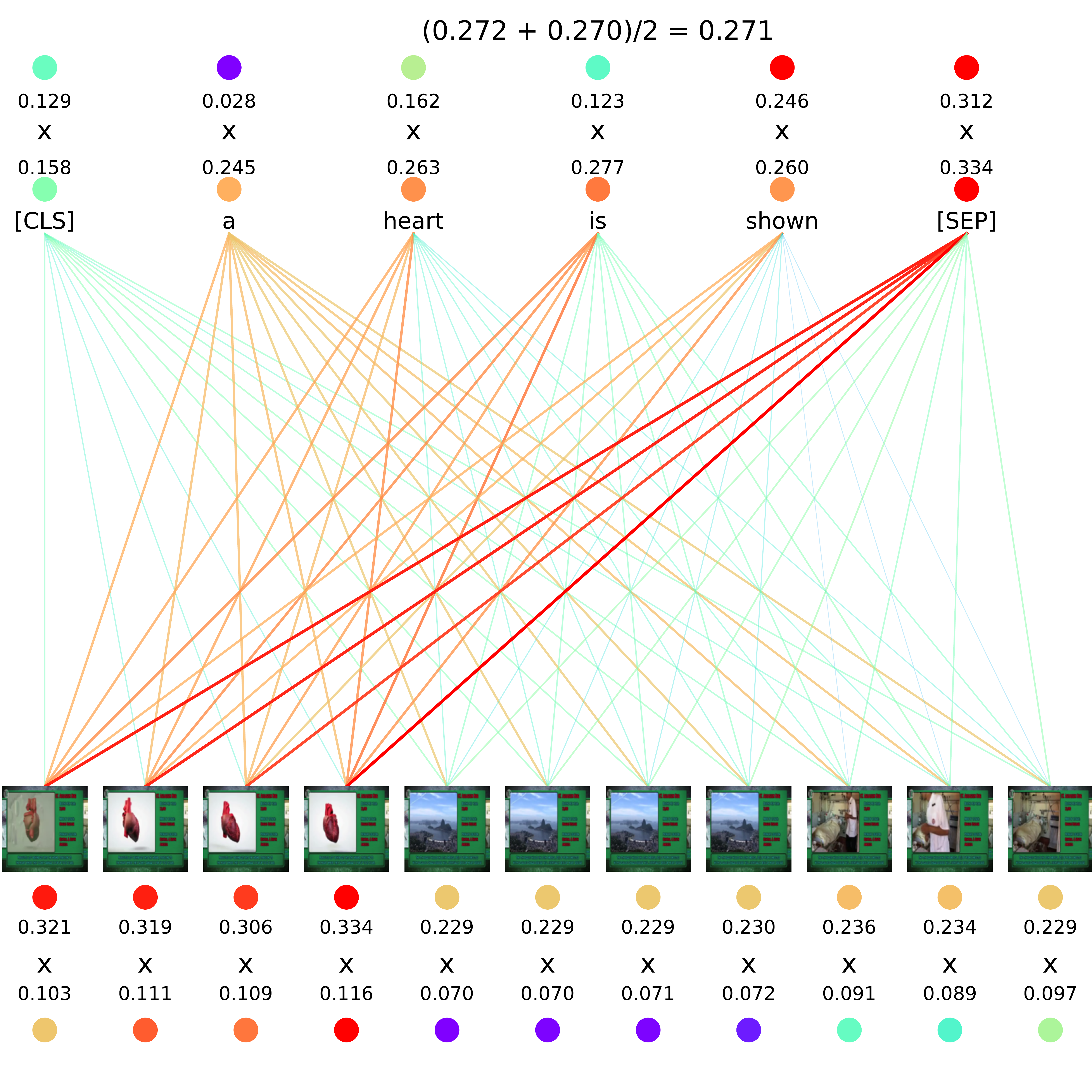}
\includegraphics[width=0.4\linewidth]{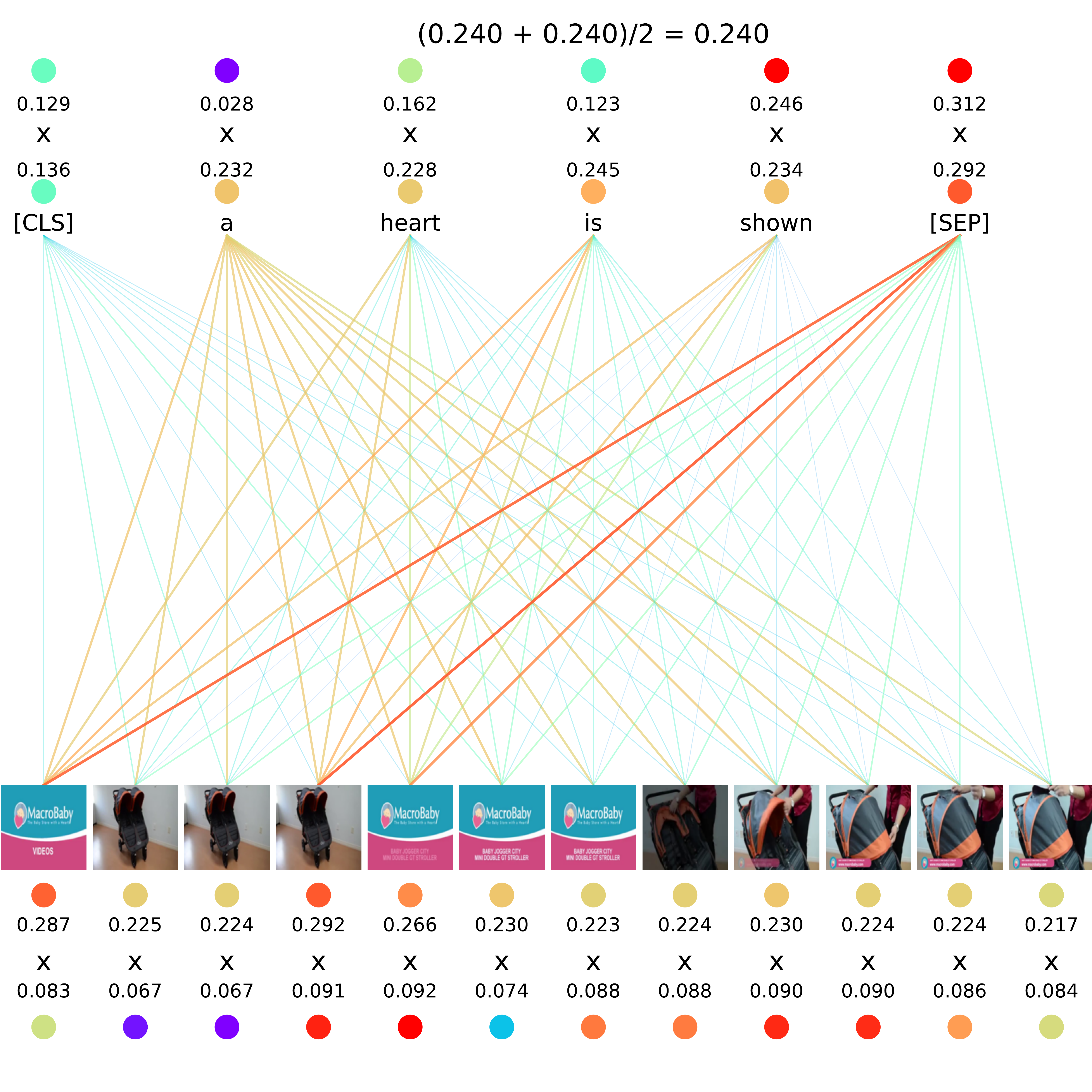}
\caption{Qualitative results on images from MSR-VTT~\cite{msrvtt}.}
\label{fig:msrvtt2}
\end{figure}

More visualization results of WTI are shown in Fig.~\ref{fig:msrvtt1} and~\ref{fig:msrvtt2}.

\clearpage
\bibliographystyle{splncs04}
\bibliography{egbib}
\end{document}